\documentclass[a4paper,fleqn]{casdc}

\usepackage[numbers]{natbib}
\usepackage{graphicx}
\usepackage{algorithmic}
\usepackage{algorithm}
\usepackage{multirow}

\usepackage{caption} 

\usepackage{algorithmic}
\usepackage{booktabs}

\usepackage{color}

\def\tsc#1{\csdef{#1}{\textsc{\lowercase{#1}}\xspace}}
\tsc{WGM}
\tsc{QE}
\tsc{EP}
\tsc{PMS}
\tsc{BEC}
\tsc{DE}

\begin{document}
\let\WriteBookmarks\relax
\def\floatpagepagefraction{1}
\def\textpagefraction{.001}
\shorttitle{Feature Concatenation Multi-view Subspace Clustering}
\shortauthors{Qinghai Zheng et~al.}

\title [mode = title]{Feature Concatenation Multi-view Subspace Clustering}                      
\author[mymainaddress]{Qinghai Zheng}[style=chinese]

\author[mymainaddress]{Jihua Zhu}[style=chinese, orcid=0000-0002-3081-8781]
\cormark[1]

\author[mymainaddress]{Zhongyu Li}[style=chinese]

\author[mymainaddress]{Shanmin Pang}[style=chinese]

\author[mysecondaddress]{Jun Wang}[style=chinese]

\author[mymainaddress]{Yaochen Li}[style=chinese]

\address[mymainaddress]{School of Software Engineering, Xi'an Jiaotong University, Xi'an 710049, People's Republic of China}

\address[mysecondaddress]{School of Communication and Information Engineering, Shanghai University, Shanghai 200444, People's Republic of China}

\cortext[cor1]{Corresponding author, email: zhujh@xjtu.edu.cn.}

\begin{abstract}
Multi-view clustering is a learning paradigm based on multi-view data. Since statistic properties of different views are diverse, even incompatible, few approaches implement multi-view clustering based on the concatenated features straightforward. However, feature concatenation is a natural way to combine multi-view data. To this end, this paper proposes a novel multi-view subspace clustering approach dubbed Feature Concatenation Multi-view Subspace Clustering (FCMSC), which boosts the clustering performance by exploring the consensus information of multi-view data. Specifically, multi-view data are concatenated into a joint representation firstly, then, $l_{2,1}$-norm is integrated into the objective function to deal with the sample-specific and cluster-specific corruptions of multiple views. Moreover, a graph regularized FCMSC is also proposed in this paper to explore both the consensus information and complementary information of multi-view data for clustering. It is noteworthy that the obtained coefficient matrix is not derived by simply applying the Low-Rank Representation (LRR) to concatenated features directly. Finally, an effective algorithm based on the Augmented Lagrangian Multiplier (ALM) is designed to optimize the objective functions. Comprehensive experiments on six real-world datasets illustrate the superiority of the proposed methods over several state-of-the-art approaches for multi-view clustering.
\end{abstract}

\begin{keywords}
Multi-view clustering\sep Subspace clustering\sep Low-rank representation \sep Feature concatenation
\end{keywords}

\maketitle

\section{Introduction}
\overfullrule=2cm
Multi-view data, which are collected from different measurements or fields to give a comprehensive description of objects, are popular in many real-world applications \cite{xu2013survey1,yin2015multi,Review1_4_Deep_multimodal,huang2018robust_nc,Review2_3_multimodal_sparse_coding,chen2019auto}. For example, in computer vision fields, an image can be presented by multiple views (GIST \cite{oliva2001modeling2}, SIFT \cite{lowe2004distinctive3}, LBP \cite{ojala2002multiresolution4}, etc.); the words presented on a webpage and the words presented in URL are two distinct views of the webpage; video signals and audio signals are two common representations and can be applied {\color{black}{to}} multimedia content understanding. Compared with single-view data, multi-view data contain both the consensus and complementary information among multiple views. And the goal of multi-view learning, which has achieved success in many applications \cite{xu2013survey1,zhao2017multi6,sun2013multi7,zhoutaomultiviewlearning2018,ZhaiLinMSC_IET,MultiviewLearning_InfSci}, is to improve the generalization performance by leveraging multiple views.

As a fundamental task in unsupervised learning, clustering, which is often used to mine underlying information of data, can be a stand-alone exploratory tool or a preprocessing step to assist other learning tasks in machine learning as well \cite{zhou2012ensemble8}. Many clustering approaches have been proposed, and subspace clustering, which assumes that high dimensional data lie in a union of low-dimensional subspaces and tries to group data points into clusters and find the corresponding subspace simultaneously, attracts lots of researches owing to its promising performance and good interpretability. In recent years, many clustering algorithms based on the subspace clustering with different constraints have been proposed \cite{hu2014smooth14,vidal2014low10,elhamifar2013sparse12,patel2015latent13,Review1_1_subspace_clustering,Review1_2_subspace_clustering,Review1_3_subspace_learning,Review2_6_clustering,Review2_5_Subspace_clustering,Review2_4_Nonnegative_matrix_clustering}. Low-Rank Subspace Clustering (LRSC) \cite{vidal2014low10} finds a low-rank linear representation of data in a dictionary of themselves and then employs the spectral clustering on an adjacent matrix, {\color{black}{which}} is derived from the low-rank representation \cite{liu2013robust11}, to obtain clustering results. Besides, Sparse Subspace Clustering (SSC) \cite{elhamifar2013sparse12}, which tries to find a sparse representation based on the $l_1$-norm, is a powerful subspace clustering algorithms as well. Additionally Low-Rank Sparse Subspace Clustering (LRSSC) \cite{patel2015latent13} applies low-rank and sparse constraints simultaneously based on the trace norm and $l_1$-norm according to the fact that the coefficient matrix is often sparse and low-rank at the same time. {\color{black}{By combining the labels and the affinity, Discriminative and Coherent Subspace Clustering (DCSC) \cite{Review1_2_subspace_clustering} tries to enhance the labels' discrimination for data in variant clusters and the affinity for data in the same cluster.}} Although these algorithms mentioned above can get promising clustering results in practice, they are designed for single-view data rather than multi-view data.

Based on the subspace clustering, many multi-view subspace clustering approaches have been proposed \cite{chao2017survey24,zhang2015low16,cao2015diversity17,MSC_InfSci,gao2015multi18,zhao2018incomplete}. Most of them process multiple views separately and obtain clustering results by finding a common shared coefficient matrix or fusing clustering results of different views directly. Although good performance has been achieved in practice, the underlying information of multi-view data is insufficiently explored in these methods. To this end, in this paper, we propose a novel multi-view subspace clustering named Feature Concatenation Multi-view Subspace Clustering (FCMSC), which performs clustering on all views simultaneously and takes advantage of the consensus information of multi-view data to improve clustering results.

{\color{black}{For multi-view clustering, a naive idea is concatenating features of all views and then running a clustering algorithm to get clustering results. Concatenated features have the following merits: 1) original information of multi-view data can be maximum preserved by concatenating features of all views into a joint view representation; 2) all views of multi-view data can be processed simultaneously during clustering. However, it is ineffective in practice and even gets worse clustering results \cite{xu2013survey1,zhao2017multi6,kumar2011co20,kumar2011co21,xia2014robust22} by simply performing a single-view clustering algorithm on the concatenated features straightforward to obtain clustering results, since each view contains its own statistical properties.}} It is noteworthy that our proposed FCMSC can achieve the promising clustering performance on the joint view representation. {\color{black}{To be specific, by introducing the concept of cluster-specific corruptions, our FCMSC decomposes the original coefficient matrix, which is derived from concatenated features by employing low-rank representation straightforward, to obtain a new low-rank coefficient matrix, which enjoys the consensus property of multi-view data.}} Moreover, a graph regularized FCMSC (gr-FCMSC) is also proposed, which can explore both the consensus information and complementary information simultaneously during clustering. Finally, an effective optimization algorithm based on the Augmented Lagrangian Multiplier (ALM) \cite{yin2015dual,lin2011linearized23} is designed for the objective functions of the proposed FCMSC and gr-FCMSC. Extensive experiments on six benchmark datasets compared with several state-of-the-arts illustrate the effectiveness and competitiveness of the proposed methods.

The main contributions of this paper can be summarized as follows:
\begin{itemize}
  \item [1)] An effective feature concatenation multi-view subspace clustering is proposed in this paper. By introducing the cluster-specific corruptions brought by different views, the proposed method can perform clustering on multiple views simultaneously and explore the consensus information of multi-view data based on the joint view representation directly.      
  \item [2)] A graph regularized feature concatenation multi-view subspace clustering (gr-FCMSC) is also proposed. By employing the graph Laplacians, both the consensus information and the complementary information of multi-view data can be fully explored during clustering.
  \item [3)] Comprehensive experiments are conducted on public available datasets, and experimental results show the effectiveness and superiority of the proposed methods compared with several state-of-the-arts. 
\end{itemize}

The rest of this paper is organized as follows. The next section reviews related works briefly. Section 3 introduces our methods, including FCMSC and gr-FCMSC, in detail. And Section 4 presents the related optimizations. Comprehensive experimental results and discussions are provided in Section 5. Finally, Section 6 provides the conclusions.

\section{Related Work}
A lot of approaches have been proposed recently to solve the multi-view clustering problem \cite{ZhouTao2019dual_multiview,zhang2017latent15,zhang2015low16,cao2015diversity17,gao2015multi18,zhao2018incomplete,kumar2011co20,kumar2011co21,xia2014robust22,GLMSC,tangchangmultiviewsubspaceclusteringTMM,chao2017survey24,yi2005multi25,tzortzis2010multiple26,wang2016multi_nc,zhao2017multi29,liu2013multi30,brbic2018MLRSSC}. Most existing multi-view clustering methods can be grouped into two main categories roughly: generative methods and discriminative methods \cite{chao2017survey24}. The idea of generative methods is to construct generative models for variant clusters respectively. For example, multi-view convex mixture models \cite{tzortzis2010multiple26} assign different weights for multiple views automatically and consider the diversity of different views. Although most generative algorithms are robust to the missing entries and have global optimization, they are accompanied with a series of hypotheses and parameters, which make the optimization more difficult and time consuming. 

Discriminative methods, the goal of which is to minimize both the intrinsic similarity of data points between different clusters and dissimilarity of data points within the same cluster through all multiple views simultaneously, have achieved good clustering results in many applications and attract the most attention of researchers in this research field \cite{chao2017survey24}. Taking examples of multi-view subspace clustering, Latent Multi-view Subspace Clustering (LMSC) \cite{zhang2017latent15} and generalized Latent Multi-view Subspace Clustering (gLMSC) \cite{GLMSC} introduce a latent representation to explore the relationships of data points among all views, obtain the underlying complementary information and seek the latent representation as well; And Multi-view Low-rank Sparse Subspace Clustering (MLRSSC) \cite{brbic2018MLRSSC} obtains multi-view clustering results by constructing an affinity matrix with the low-rank and sparsity constraints; Multi-view subspace clustering by learning a joint affinity graph \cite{tangchangmultiviewsubspaceclusteringTMM} pursuits a low-rank subspace representation with diversity regularization and a rank constraint for multi-view clustering. Besides, many spectral clustering based methods are also proposed in recent years. The co-training approach for multi-view spectral clustering \cite{kumar2011co20} and the co-regularized multi-view spectral clustering \cite{kumar2011co21} try to get clustering results that can maximize the similarity graph agreement among different views; Robust Multi-view Spectral Clustering (RMSC) \cite{xia2014robust22} recovers a common transition probability matrix via low-rank and sparse decomposition and employs the Markov chain approach to obtain clustering results. In addition, some multi-view clustering methods based on the matrix factorization method \cite{lee1999learning31} are proposed by exploring the consensus information among views \cite{zhao2017multi29,liu2013multi30}. For most of discriminative multi-view clustering methods, the essential difference is the style they use to explore the underlying information of multiple views.

{\color{black}{Inspire by the success of deep learning \cite{lecun2015deep,Review1_5_Deep_Hierachical,Review2_2_unsupervised_dimension_reduction,Review2_1_Triplet_Loss_Place_Recognition}, some clustering methods based on deep learning are proposed \cite{DeepSC,DASC}. By introducing a self-expressive layer, Deep Subspace Clustering Networks can map data to a latent space non-linearly and learns the affinity matrix straightforward. Deep Canonical Correlation Analysis (DCCA) \cite{DCCA} and Deep Canonically Correlated Autoencoders (DCCAE) \cite{DCCAE} are two deep learning based methods which can be employed for multi-view clustering.}} Although good clustering results can be achieved, most existing multi-view clustering methods deal with different views separately, and that is an ineffective way since the relationships among multiple views are ignored. A natural way is to combine all views before clustering, and some related approaches have been proposed \cite{chaudhuri2009multi32,chao2016multi33,zhang2017latent15,zhang2006linear34,Guo2014Multiple35}. However, these methods may corrupt either the consensus information or the complementary information among views during combination to varying degrees. Taking the joint view representation into consideration, multi-view clustering results achieved by employing a single-view clustering algorithm to the joint view representation directly are uncompetitive \cite{zhang2015low16,cao2015diversity17,kumar2011co21,xia2014robust22,chao2017survey24}, and few works focus on this kind of combination styles. However, it is obvious that original information contained among multiple views can get maximum preservation by concatenating features of all views straightforward. It is notable that the proposed FCMSC and gr-FCMSC can get promising and competitive clustering results by utilizing the concatenated features of multiple views straightforward.

\begin{figure*}
\center
\includegraphics[width=1\linewidth]{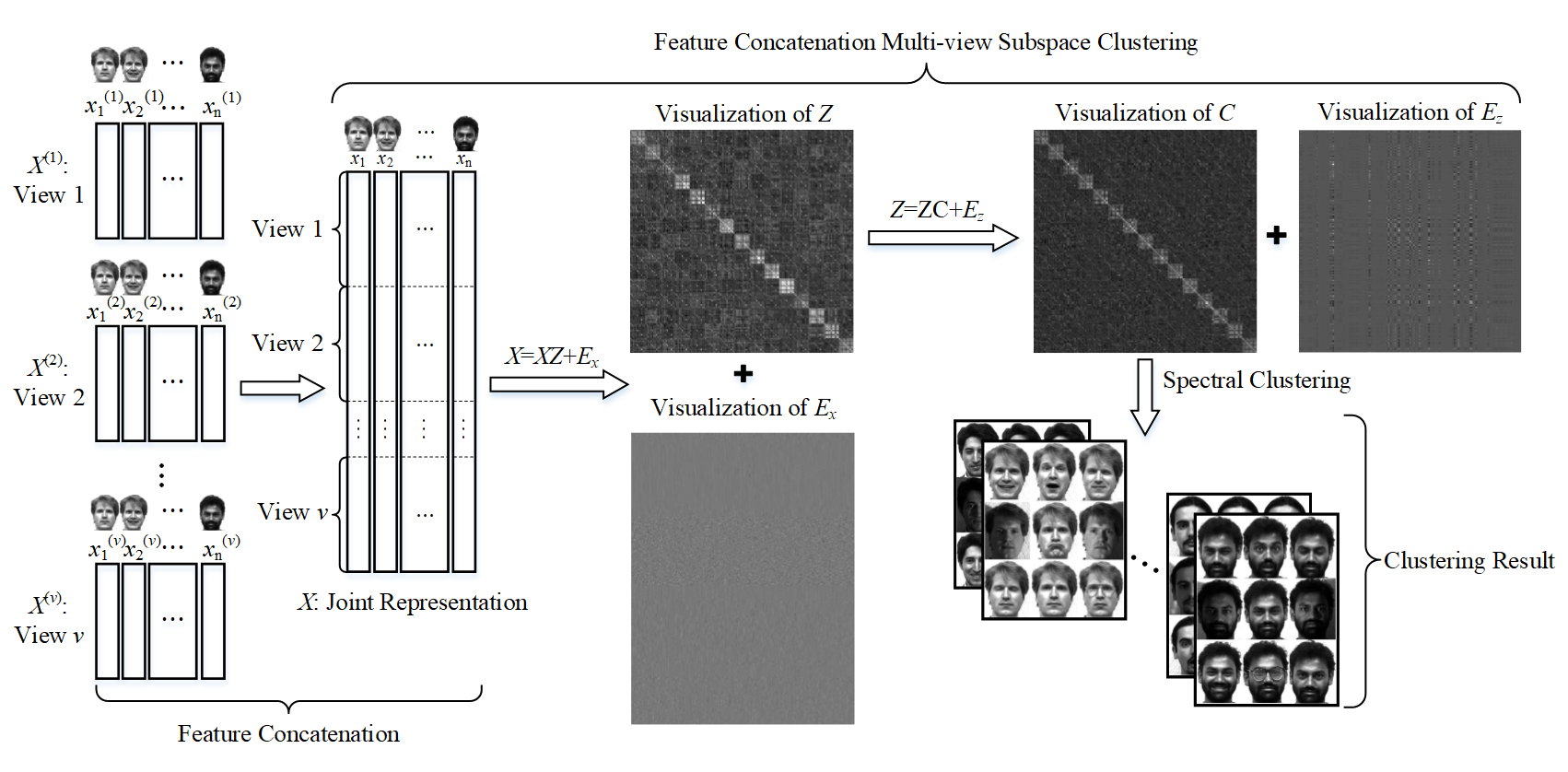}
\caption{Illustration of the proposed method, in which multi-view subspace clustering is implemented on the joint view representation. Multiple views are concatenated firstly, then the FCMSC is employed to obtain a desired coefficient matrix, i.e. $C$, and the last step is to infer the clustering results of data points by leveraging the spectral clustering approach with the adjacency matrix 
$({\mathop{\rm abs}\nolimits} (C) + {\mathop{\rm abs}\nolimits} ({C^T}))/2$.}
\center
\label{Section3_FCMSC}
\end{figure*}

\section{Feature Concatenation Multi-view Subspace Clustering}
In this section, we propose the Feature Concatenation Multi-view Subspace Clustering (FCMSC) method by exploring the consensus information of multi-view data. Moreover, a graph regularized FCMSC method, termed as gr-FCMSC, is also proposed, and it can explore both the consensus information and complementary information of multiple views for multi-view clustering.

\subsection{FCMSC}
For convenience, Table \ref{table_symbols} lists main symbols leveraged throughout this paper. Given a multi-view dataset with $v$ views and $n$ samples, i.e. $\{ x_1^{(i)},x_2^{(i)}, \cdots ,x_n^{(i)}\} _{i = 1}^v$, data points of which are drawn from $m$ multiple subspaces. In order to obtain a matrix that each column has the same magnitude, data of each view are normalized within the range of $\left[ {0,1} \right]$, and then multiple views are concatenated into a joint view representation matrix $X$, which is defined as follows:
\begin{equation}
X = \left[ {\begin{array}{*{20}{c}}
{x_1^{(1)}}&{x_2^{(1)}}& \cdots &{x_n^{(1)}}\\
{x_1^{(2)}}&{x_2^{(2)}}& \cdots &{x_n^{(2)}}\\
 \vdots & \vdots & \ddots & \vdots \\
{x_1^{(v)}}&{x_2^{(v)}}& \cdots &{x_n^{(v)}}
\end{array}} \right],
\end{equation}
where $x_i^{(k)}$ denotes the features of the $i$-th sample from the $k$-th view, and the $i$-th column of $X$ contains features of all views of the $i$-th sample. Based on the concatenated features, Fig. \ref{Section3_FCMSC} displays the framework of the proposed FCMSC.

\begin{table}
\renewcommand{\arraystretch}{1.3}
\caption{Main Symbols}
\label{table_symbols}
\centering
\begin{tabular}{l|l}
\toprule
\bfseries Symbol & \bfseries Meaning\\
\midrule
$n$ & The number of samples.\\
$v$ & The number of views.\\
$m$ & The number of clusters.\\
$d_i$ & The dimension of features in $i$-th view.\\
$d$ & The dimension of the concatenated features.\\
$x_k^{(i)} \in {R^{{d_i}}}$ & The features of $k$-th sample from $i$-th view.\\
$X \in {R^{d \times n}}$ & The joint view representation matrix.\\
$Z \in {R^{n \times n}}$ & The original coefficient matrix.\\
${E_x} \in {R^{d \times n}}$ & The sample-specific corruptions.\\
$C \in {R^{n \times n}}$ & The desired coefficient matrix.\\
${E_{cs}} \in {R^{d \times n}}$ & The cluster-specific corruptions.\\
${E_z} \in {R^{n \times n}}$ & The term derived from ${E_{cs}}$.\\
${L_i} \in {R^{n \times n}}$ & The Laplacian matrix of $i$-th view.\\
${\left\| A \right\|_*}$ & The trace-norm of matrix $A$.\\
${\left\| A \right\|_{2,1}}$ & The $l_{2,1}$-norm of matrix $A$.\\
\bottomrule
\end{tabular}
\end{table}

Since statistic properties of different views are diverse, even incompatible among views, it is difficult to explore the mutual information of multiple views effectively and fully. In order to get a preliminary exploration of multi-view data, we consider the following objective function in the beginning:
\begin{equation}
\begin{array}{l}
\mathop {\min }\limits_{Z,{E_x}} {\left\| {{E_x}} \right\|_{2,1}} + \lambda {\left\| Z \right\|_*}\\
{\rm{s}}{\rm{.t}}{\rm{.}}{\kern 1pt} {\kern 1pt} {\kern 1pt}{\kern 1pt} {\kern 1pt} {\kern 1pt} X = XZ + {E_x},
\end{array}
\label{StandLRR}
\end{equation}
where $Z$ indicates an original coefficient matrix of $X$, $E_x$ denotes the sample-specific corruptions of data points, and $\lambda$ is a trade-off parameter. The $l_{2,1}$-norm of $E_x$ enforces $E_x$ to be sparse in columns and columns of $E_x$ to be zero. Equation (\ref{StandLRR}) is a standard low-rank representation \cite{liu2013robust11} of the concatenated features. However, experimental results presented in \cite{zhang2015low16,cao2015diversity17,chao2017survey24} and later section of this paper show that the clustering performance is uncompetitive if a spectral clustering algorithm is performed based on the coefficient matrix $Z$. This is because each view has specific statistical properties, which may be contradictory among views, and it is unreasonable to explore the joint views representation by directly employing single-view clustering algorithm.

\begin{figure}
\center
\includegraphics[width=3.3in]{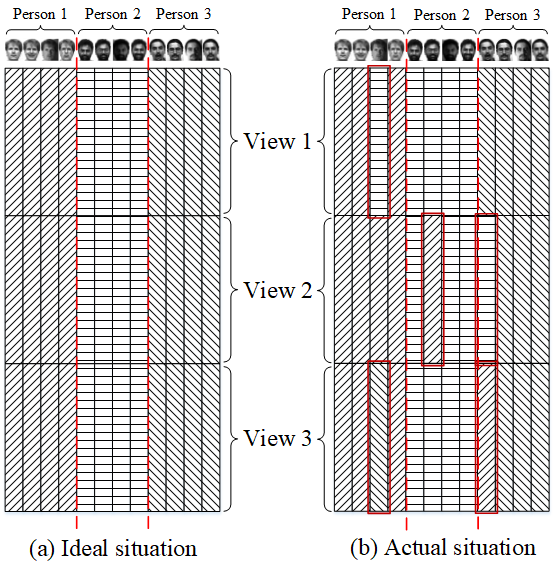}
\caption{Illustrating the cluster-specific corruptions brought by multiple views. Taking 12 images of 3 individuals for examples, and they are described by 3 views. With the joint view representation, (a) is an ideal situation that clustering results of all views are coincident. (b) is the actual situation since clustering results of different views are different to some degree. The columns of joint view representation matrix, containing red rectangles, indicate the cluster-specific corruptions, which are brought by multiple views obviously.}
\center
\label{Section3_ClusterSpecific}
\end{figure}

In this paper, we introduce the cluster-specific corruptions, which are accompanied with multi-view data, as shown in Fig. \ref{Section3_ClusterSpecific}. Without considering the cluster-specific corruptions, it is expected that running a single-view clustering algorithm on the concatenated features is hard to get satisfied clustering results. And the original coefficient matrix $Z$, obtained in (\ref{StandLRR}), is far from good enough for multi-view clustering. In order to handle the concatenated features better and get a desired coefficient matrix, it is suggested to consider the following formulation:
\begin{equation}
X = DC + {E_{cs}} + {E_x},
\label{FCMSC_equation_naive}
\end{equation}
where $D$ indicates a dictionary matrix, $C$ denotes the desired coefficient matrix, and $E_{cs}$ represents the cluster-specific corruptions among multiple views. Equation (\ref{FCMSC_equation_naive}) considers both the cluster-specific and sample-specific corruptions.

Obviously, the choices of $D$ and ${E_{cs}}$ are vital for the final multi-view clustering performance. Since matrix $D$ is free of the sample-specific corruptions, it is reasonable to employ the reconstructed features, obtained from (\ref{StandLRR}), as the dictionary matrix, i.e. $D=XZ$. For ${E_{cs}}$, most existing norms are not suitable for it. As shown in Fig. \ref{Section3_ClusterSpecific}, under the assumption that the true underlying clustering would assign corresponding points in each view to the same cluster, the number of columns with cluster-specific corruptions in matrix $X$ should be small, and the major part of columns {\color{black}{achieves}} the same clustering results. However, it is {\color{black}{difficult}} to process the cluster-specific corruptions directly. {\color{black}{In this paper, we decomposed $E_{cs}$ as $E_{cs} = XE_z$. By assuming that $E_z$ is sparse in columns, $XE_z$ can capture the cluster-specific corruption approximately. So it is reasonable to impose the $l_{2,1}$-norm minimization constraint on $E_z$, and $E_{cs} = XE_z$ can characterize the cluster-specific corruptions of multi-view data properly.}} Accordingly, (\ref{FCMSC_equation_naive}) is rewritten as follows:
\begin{equation}
X = XZC + X{E_z} + {E_x}.
\label{FCMSC_equation_proposed}
\end{equation}

For simplicity, we can reformulate the above equation as follows:
\begin{equation}
X = X(ZC + {E_z}) + {E_x}.
\end{equation}

\begin{figure}
\center
\includegraphics[width=3.2in]{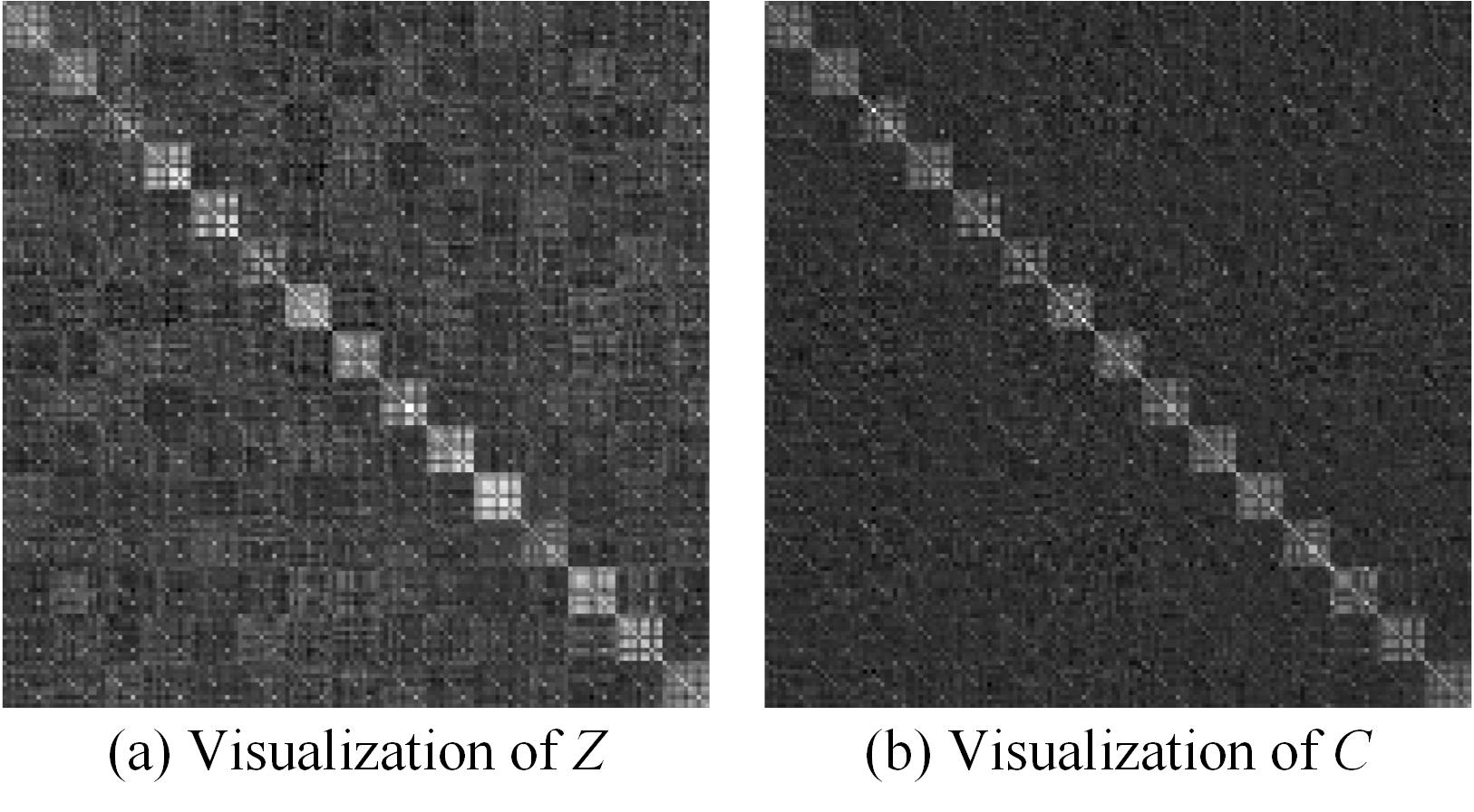}
\caption{Visualization of coefficient matrices obtained from the Yale Face dataset. (a) is derived from (\ref{StandLRR}), and (b) is calculated from (\ref{FCMSC_ObjectiveFunction}). where $Z$ and $C$ are both displayed in the form of (abs($Z^T$)+abs($Z$))/2 and (abs($C^T$)+abs($C$))/2, and obviously, $C$ characters the underlying clustering structures of data much better than $Z$.}
\center
\label{Section3_Z_C}
\end{figure}

As a consequence, it is straightforward to design the following objective function for multi-view clustering based on the joint view representation $X$:
\begin{equation}
\begin{array}{*{20}{l}}
{\mathop {\min }\limits_{Z,C,{E_x},{E_z}} {{\left\| {{E_x}} \right\|}_{2,1}} + {\lambda _1}{{\left\| {{E_z}} \right\|}_{2,1}} + {\lambda _2}{{\left\| C \right\|}_*}}\\
{{\rm{s}}{\rm{.t}}{\rm{.}}{\kern 1pt} {\kern 1pt} {\kern 1pt}{\kern 1pt} {\kern 1pt} {\kern 1pt} X = XZ + {E_x},{\kern 1pt} {\kern 1pt} {\kern 1pt} Z = ZC + {E_z},}
\end{array}
\label{FCMSC_ObjectiveFunction}
\end{equation}
where $\lambda_1$ and $\lambda_2$ are trade-off parameters. Although $E_x$ and $E_z$ are both imposed with the $l_{2,1}$-norm constraint, they are totally different in essence. {\color{black}{More specifically, $E_x$ illustrates the sample-specific corruptions, and $E_{cs} = XE_z$ is used to process the cluster-specific corruptions caused by multiple-views.}} Theoretically, compared with the coefficient matrix obtained in (\ref{StandLRR}), the coefficient matrix $C$ is much better for multi-view clustering. To view the difference in a more intuitive way, Fig. \ref{Section3_Z_C} displays a visualization of $Z$ and $C$ conducted on the Yale Face dateset\footnote[1]{The Yale Face database contains 165 grayscale images in GIF format of 15 individuals. More details will be presented in  the section of experiment.}. As shown in Fig. \ref{Section3_Z_C}, it is clear that the matrix $C$ has more suitable structures than $Z$ for clustering.

\subsection{gr-FCMSC}
In FCMSC, only the consensus information of multi-view data is employed for clustering. In order to leverage the complementary information as well, a graph regularized Feature Concatenated Multi-view Subspace Clustering (gr-FCMSC) is also proposed in this paper. Graph regularization can preserve local manifold structures \cite{cai2010graph,xia2010MSE_graphregularized,gao2012laplacian}, and inspired by \cite{xia2010MSE_graphregularized} we impose the following graph Laplacian regularizer of multiple views on the FCMSC to explore the complementary of multi-view data:
\begin{equation}
\sum\limits_{i = 1}^v {Tr({C^T}{L_i}C)},
\end{equation}
where $C^T$ denotes the transpose of $C$, $L_i$ represents the graph Laplacian matrix of $i$-th view, $L_i=D_i-W_i$, and $D_i$ is the degree matrix of the $i$-th view, $W_i$ is the adjacency matrix of the $i$-th view \cite{Luxburg2007A37}. And, the objective function of gr-FCMSC can be formulated as follows:
\begin{equation}
\begin{array}{l}
\mathop {\min }\limits_{Z,C,{E_x},{E_z}} {\left\| {{E_x}} \right\|_{2,1}} + {\lambda _1}{\left\| {{E_z}} \right\|_{2,1}} + {\lambda _2}{\left\| C \right\|_ * }\\
{\kern 1pt} {\kern 1pt} {\kern 1pt} {\kern 1pt} {\kern 1pt} {\kern 1pt} {\kern 1pt} {\kern 1pt} {\kern 1pt} {\kern 1pt} {\kern 1pt} {\kern 1pt} {\kern 1pt} {\kern 1pt} {\kern 1pt} {\kern 1pt} {\kern 1pt} {\kern 1pt} {\kern 1pt} {\kern 1pt} {\kern 1pt} {\kern 1pt} {\kern 1pt} {\kern 1pt} {\kern 1pt} {\kern 1pt} {\kern 1pt} {\kern 1pt} {\kern 1pt} {\kern 1pt} {\kern 1pt} {\kern 1pt}{\kern 1pt} {\kern 1pt} {\kern 1pt} {\kern 1pt} {\kern 1pt}{\kern 1pt} {\kern 1pt} + {\lambda _3}\sum\limits_{i = 1}^v {Tr({C^T}{L_i}C)} \\
{\rm{s}}{\rm{.t}}{\rm{.}}{\kern 1pt} {\kern 1pt} {\kern 1pt}{\kern 1pt} {\kern 1pt} {\kern 1pt} X = XZ + {E_x},{\kern 1pt} {\kern 1pt} {\kern 1pt} Z = ZC + {E_z}
\end{array}
\label{grFCMSC_ObjectiveFunction}
\end{equation}
where ${\lambda _1}$, ${\lambda _2}$, and ${\lambda _3}$ denote trade-off parameters. Obviously, the desired coefficient matrix $C$, derived from (\ref{grFCMSC_ObjectiveFunction}), takes specific manifold structures of different views into consideration, in other words, the complementary information of multi-view data are also leveraged for clustering.

Once the desired coefficient matrix $C$ is learned, we construct an adjacency matrix for spectral clustering to get multi-view clustering results as follows:
\begin{equation}
W = \frac{{{\mathop{\rm abs}\nolimits} (C) + {\mathop{\rm abs}\nolimits} ({C^T})}}{2},
\end{equation}
where abs($ \cdot $) denotes the absolution function, which can deal with a matrix and return the absolute value of each element in the matrix.

\section{Optimization}
In this section, the optimization algorithms of the objective functions, including FCMSC and gr-FCMSC, are introduced in detail, then the computational complexity and convergence are analyzed as well.

\begin{algorithm}[t]
\caption{Optimization of the proposed FCMSC}
\label{Algorithm1}
{\bf INPUT:} \\
\hspace*{0.06in} Multi-view data $\{ x_1^{(i)},x_2^{(i)}, \cdots ,x_n^{(i)}\} _{i = 1}^v$;\\
\hspace*{0.06in} $E_x=0$, $E_z=0$, $C=0$, $J=0$,\\
\hspace*{0.06in} $Y_1=0$, $Y_2=0$, $Y_3=0$,\\
\hspace*{0.06in} $\mu  = {10^{ - 4}}$, ${\mu _{\max }} = {10^{6}}$, $\varepsilon  = {10^{ - 6}}$,\\
\hspace*{0.06in} Initialize Z with random values;\\
{\bf OUTPUT:} \\
\hspace*{0.06in} $C$, $Z$, $E_z$, $E_x$; 

\MakeUppercase{\algorithmicrepeat} \\
\hspace*{0.06in} {\color{black}{Update $E_x$ according to the problem (12);}}\\
\hspace*{0.06in} {\color{black}{Update $E_z$ according to the problem (14);}}\\
\hspace*{0.06in} {\color{black}{Update $J$ according to the problem (15);}}\\
\hspace*{0.06in} {\color{black}{Update $C$ according to the problem (18);}}\\
\hspace*{0.06in} {\color{black}{Update $Z$ according to the problem (20);}}\\
\hspace*{0.06in} {\color{black}{Update $Y_1$, $Y_2$, $Y_3$ and $\mu$ according to the problem (23);}}\\
\MakeUppercase{\algorithmicuntil}\\
\hspace*{0.06in} ${\left\| {X - XZ - {E_x}} \right\|_\infty } < \varepsilon, $\\
\hspace*{0.06in} ${\left\| {Z - ZC - {E_z}} \right\|_\infty } < \varepsilon, $\\
\hspace*{0.06in} ${\rm{and}}{\kern 1pt} {\kern 1pt} {\kern 1pt} {\kern 1pt} {\left\| {C - J} \right\|_\infty } < \varepsilon.$
\end{algorithm}

\subsection{Optimization for FCMSC}
Although the optimization problem of (\ref{FCMSC_ObjectiveFunction}) is not convex with respect to the variables, i.e. $Z$, $C$, $E_x$, and $E_z$, jointly, subproblems with respect to each of them are convex. So we apply the Alternating Direction Minimization strategy based on the Augmented Lagrangian Multiplier (ALM) \cite{lin2011linearized23} method to solve the objective function (\ref{FCMSC_ObjectiveFunction}) effectively. Additionally, an auxiliary variable is introduced here so as to make the objective function separable and convenient for optimization. Accordingly, (\ref{FCMSC_ObjectiveFunction}) can be reformulated equivalently as follows:

\begin{equation}
\begin{array}{l}
\mathop {\min }\limits_{Z,C,{E_z},{E_x},J} {\left\| {{E_x}} \right\|_{2,1}} + {\lambda _1}{\left\| {{E_z}} \right\|_{2,1}} + {\lambda _2}{\left\| J \right\|_ * }\\
{\rm{s}}{\rm{.t}}{\rm{.}}{\kern 1pt} {\kern 1pt} {\kern 1pt} {\kern 1pt} {\kern 1pt} {\kern 1pt} X = XZ + {E_x},{\kern 1pt} {\kern 1pt} {\kern 1pt} Z = ZC + {E_z},{\kern 1pt} {\kern 1pt} {\kern 1pt} C = J.
\end{array}
\label{FCMSC_ObjectiveFunction_withJ}
\end{equation}
where $J$ denotes the auxiliary variable, ${\lambda _1}$ and ${\lambda _2}$ are tradeoff parameters.

The corresponding ALM problem of (\ref{FCMSC_ObjectiveFunction_withJ}), which should be minimized in this section,can be shown as follows:
\begin{equation}
\displaystyle
\begin{array}{l}
{\cal L}({E_x},{E_z},J,C,{Y_1},{Y_2},{Y_3},\mu)\\
 {\kern 1pt} {\kern 1pt} {\kern 1pt}= {\left\| {{E_x}} \right\|_{2,1}} + {\lambda _1}{\left\| {{E_z}} \right\|_{2,1}} + {\lambda _2}{\left\| J \right\|_ * }\\
 {\kern 1pt} {\kern 1pt} {\kern 1pt}+ \left\langle {{Y_1},X - XZ - {E_x}} \right\rangle  + \frac{\mu }{2}\left\| {X - XZ - {E_x}} \right\|_F^2\\
 {\kern 1pt} {\kern 1pt} {\kern 1pt}+ \left\langle {{Y_2},Z - ZC - {E_z}} \right\rangle  + \frac{\mu }{2}\left\| {Z - ZC - {E_z}} \right\|_F^2\\
 {\kern 1pt} {\kern 1pt} {\kern 1pt}+ \left\langle {{Y_3},C - J} \right\rangle  + \frac{\mu }{2}\left\| {C - J} \right\|_F^2,
\end{array}
\label{ALM_FCMSC_ObjectiveFunctin}
\end{equation}
where $Y_1$, $Y_2$, and $Y_3$ are Laplacian multipliers, $\mu$ indicates a positive adaptive penalty parameter, $\left\langle {A,B} \right\rangle$  denotes the trace of ${A^T}B$.

Since the Alternating Direction Minimization strategy is employed to minimize the above ALM problem, the whole problem is decomposed into several subproblems, which are convex and can be optimized effectively.

\textbf{1) Updating $E_x$:} To update $E_x$ with other variables fixed, the following minimization problem should be optimized:
\begin{equation}
\mathop {\min }\limits_{{E_x}} {\left\| {{E_x}} \right\|_{2,1}}{\rm{ + }}\frac{\mu }{2}\left\| {{E_x} - (X - XZ + \frac{{{Y_1}}}{\mu })} \right\|_F^2,
\end{equation}
which has a closed-form solution.

Specifically, the solution of the above subproblem is denoted as $E_x^*$, and we can get the following closed-form solution \cite{yang2009fast}:
\begin{equation}
{\left[ {E_x^*} \right]_{:,j}} = \left\{ \begin{array}{l}
\frac{{{{\left\| {{{\left[ {{T_E}} \right]}_{:,j}}} \right\|}_2} - \frac{1}{\mu }}}{{{{\left\| {{{\left[ {{T_E}} \right]}_{:,j}}} \right\|}_2}}}{\left[ {{T_E}} \right]_{:,j}},{\kern 1pt} {\kern 1pt} {\kern 1pt} {\rm{if}}{\kern 1pt} {\kern 1pt} {\kern 1pt} {\left\| {{{\left[ {{T_E}} \right]}_{:,j}}} \right\|_2} > \frac{1}{\mu }\\
{\kern 1pt} {\kern 1pt} {\kern 1pt} {\kern 1pt} {\kern 1pt} {\kern 1pt} {\kern 1pt} {\kern 1pt} {\kern 1pt} {\kern 1pt} {\kern 1pt} {\kern 1pt} {\kern 1pt} {\kern 1pt} {\kern 1pt} {\kern 1pt} {\kern 1pt} {\kern 1pt} {\kern 1pt} {\kern 1pt} {\kern 1pt} {\kern 1pt} {\kern 1pt} {\kern 1pt} {\kern 1pt} {\kern 1pt} {\kern 1pt} {\kern 1pt} {\kern 1pt} {\kern 1pt} {\kern 1pt} {\kern 1pt} {\kern 1pt} {\kern 1pt} {\kern 1pt} {\kern 1pt} {\kern 1pt} {\kern 1pt} {\kern 1pt} {\kern 1pt} {\kern 1pt} {\kern 1pt} {\kern 1pt} 0{\kern 1pt} {\kern 1pt} {\kern 1pt} {\kern 1pt} {\kern 1pt} {\kern 1pt} {\kern 1pt} {\kern 1pt} {\kern 1pt} {\kern 1pt} {\kern 1pt} {\kern 1pt} {\kern 1pt} {\kern 1pt} {\kern 1pt} {\kern 1pt} {\kern 1pt} {\kern 1pt} {\kern 1pt} {\kern 1pt} {\kern 1pt} {\kern 1pt} {\kern 1pt} {\kern 1pt} {\kern 1pt} {\kern 1pt} {\kern 1pt} {\kern 1pt} {\kern 1pt} {\kern 1pt} {\kern 1pt} {\kern 1pt}  ,{\kern 1pt} {\kern 1pt} {\kern 1pt} {\rm{otherwise}}
\end{array} \right.,
\end{equation}
where ${\left[ A \right]_{:,j}}$ represents the $j$-th column of the matrix $A$, and ${T_E} = X - XZ + \frac{{{Y_1}}}{\mu }$.

\textbf{2) Updating $E_z$:} The subproblem of updating $E_z$, in which other variables are all fixed, can be written as follows:
\begin{equation}
\mathop {\min }\limits_{{E_z}} {{\lambda_1}\left\| {{E_z}} \right\|_{2,1}} + \frac{\mu }{2}\left\| {{E_z} - (Z - ZC + \frac{{{Y_2}}}{\mu })} \right\|_F^2.
\end{equation}

This subproblem is similar to the subproblem of updating $E_x$, and can be optimized effectively in the same way.

\textbf{3) Updating $J$:} With other variables fixed, we solve the following problem to update variable $J$:
\begin{equation}
\mathop {\min }\limits_J {\lambda _2}{\left\| J \right\|_*} + \frac{\mu }{2}\left\| {J-(C + \frac{{{Y_3}}}{\mu })} \right\|_F^2,
\end{equation}
which can be optimized by leveraging the singular value threshold method \cite{lin2011linearized23}. Specifically, by setting ${T_J} = C + {Y_3}/\mu $ and performing singular value decomposition (SVD) on $T_J$, i.e. ${T_J} = U\Sigma {V^T}$, we achieve the optimization as follows:
\begin{equation}
J = U{S_{{\lambda _2}/\mu }}(\Sigma ){V^T},
\end{equation}
where ${S_\varepsilon }$  denotes a soft-thresholding operator as following and can be extended to matrices by applying it element-wise.
\begin{equation}
{S_\varepsilon }(x) = \left\{ \begin{array}{l}
x - \varepsilon ,{\kern 1pt} {\kern 1pt} {\kern 1pt} {\rm{if}}{\kern 1pt} {\kern 1pt} {\kern 1pt} x - \varepsilon  > 0\\
x + \varepsilon ,{\kern 1pt} {\kern 1pt} {\kern 1pt} {\rm{if}}{\kern 1pt} {\kern 1pt} {\kern 1pt} x - \varepsilon  < 0\\
{\kern 1pt} {\kern 1pt} {\kern 1pt} {\kern 1pt} {\kern 1pt} {\kern 1pt} {\kern 1pt} {\kern 1pt} 0{\kern 1pt} {\kern 1pt} {\kern 1pt} {\kern 1pt} {\kern 1pt} {\kern 1pt} {\kern 1pt} {\kern 1pt} {\kern 1pt} {\kern 1pt} ,{\kern 1pt} {\kern 1pt} {\kern 1pt} {\rm{otherwise}}.
\end{array} \right.
\end{equation}

\textbf{4) Updating $C$:} When other variables are fixed, the subproblem with respect to $C$ can be written as follows:
\begin{equation}
\begin{array}{l}
\mathop {\min }\limits_C \left\langle {{Y_2},Z - ZC - {E_z}} \right\rangle  + \frac{\mu }{2}\left\| {Z - ZC - {E_z}} \right\|_F^2\\
{\kern 1pt} {\kern 1pt} {\kern 1pt} {\kern 1pt} {\kern 1pt} {\kern 1pt} {\kern 1pt} {\kern 1pt} {\kern 1pt} {\kern 1pt} {\kern 1pt} {\kern 1pt} {\kern 1pt} {\kern 1pt} {\kern 1pt} {\kern 1pt} {\kern 1pt} {\kern 1pt} {\kern 1pt} {\kern 1pt} {\kern 1pt} {\kern 1pt} {\kern 1pt} {\kern 1pt} {\kern 1pt} {\kern 1pt} {\kern 1pt} {\kern 1pt} {\kern 1pt} {\kern 1pt} {\kern 1pt} {\kern 1pt} {\kern 1pt} {\kern 1pt} {\kern 1pt} {\kern 1pt} {\kern 1pt} {\kern 1pt} {\kern 1pt}  + \left\langle {{Y_3},C - J} \right\rangle  + \frac{\mu }{2}\left\| {C - J} \right\|_F^2.
\end{array}
\label{FCMSC_Subproblem_C}
\end{equation}

In order to get an optimization, we take the derivative of the above function with respect to variable $C$ and let the derivative to be zero, then obtain the following solution:
\begin{equation}
\begin{array}{l}
C = T_{CA}^{ - 1}{T_{CB}},\\
{T_{CA}} = \mu (I + {Z^T}Z),\\
{T_{CB}} = \mu J - {Y_3} + {Z^T}{Y_2} + \mu ({Z^T}Z - {Z^T}{E_z}),
\end{array}
\end{equation}
where $I$ is an identity matrix with the proper size.

\textbf{5) Updating $Z$:} With other variables being fixed, the subproblem of updating $Z$ can be written as follows:
\begin{equation}
\begin{array}{l}
\mathop {\min }\limits_Z \left\langle {{Y_1},X - XZ - {E_x}} \right\rangle  + \frac{\mu }{2}\left\| {X - XZ - {E_x}} \right\|_F^2\\
{\kern 1pt} {\kern 1pt} {\kern 1pt} {\kern 1pt} {\kern 1pt} {\kern 1pt} {\kern 1pt} {\kern 1pt} {\kern 1pt}  + \left\langle {{Y_2},Z - ZC - {E_z}} \right\rangle  + \frac{\mu }{2}\left\| {Z - ZC - {E_z}} \right\|_F^2.
\end{array}
\label{FCMSC_Subproblem_Z_1}
\end{equation}

Differentiating (\ref{FCMSC_Subproblem_Z_1}) with respect to Z and letting it to be zero, the following equivalent equation can be achieved, solution of which is the optimization of this subproblem:
\begin{equation}
{T_{ZA}}Z + Z{T_{ZB}} = {T_{ZC}},
\label{FCMSC_Subproblem_Z_2}
\end{equation}
where $T_{ZA}$, $T_{ZB}$, and $T_{ZC}$ can be written as follows:
\begin{equation}
\begin{array}{l}
{T_{ZA}} = {X^T}X + I,\\
{T_{ZB}} = C{C^T} - C - {C^T},\\
{T_{ZC}} = {X^T}X - {X^T}{E_x} + {E_z} - {E_z}{C^T}\\
{\kern 1pt} {\kern 1pt} {\kern 1pt} {\kern 1pt} {\kern 1pt} {\kern 1pt} {\kern 1pt} {\kern 1pt} {\kern 1pt} {\kern 1pt} {\kern 1pt} {\kern 1pt} {\kern 1pt} {\kern 1pt} {\kern 1pt} {\kern 1pt} {\kern 1pt} {\kern 1pt} {\kern 1pt} {\kern 1pt} {\kern 1pt} {\kern 1pt} {\kern 1pt} {\kern 1pt} {\kern 1pt} {\kern 1pt} {\kern 1pt} {\rm{ + }}\frac{1}{\mu }{X^T}{Y_1} + \frac{1}{\mu }({Y_2}{C^T} - {Y_2}).
\end{array}
\end{equation}

Equation (\ref{FCMSC_Subproblem_Z_2}) is a Sylvester equation and can be optimized effectively referring to \cite{Bartels1972Solution38}.

\textbf{6) Updating Lagrange multipliers and $\mu$:} According to \cite{lin2011linearized23}, we update the Lagrange multipliers and the parameter $\mu $ as following:
\begin{equation}
\begin{array}{l}
{Y_1} = {Y_1} + \mu (X - XZ - {E_x}),\\
{Y_2} = {Y_2} + \mu (Z - ZC - {E_z}),\\
{Y_3} = {Y_3} + \mu (C - J),\\
\mu  {\kern 1pt} {\kern 1pt} {\kern 1pt}{\kern 1pt} = \min (\rho \mu ,{\mu _{\max }}),
\end{array}
\end{equation}
where $\rho>1$ and the parameter $\mu$ is monotonically increased by $\rho$ until reaching the maximum, ${\mu _{\max }}$.

Algorithm 1 outlines the whole procedure of optimization for FCMSC. It is should be noticed that we random initialize $Z$ in practice to avoid all zeros solutions.

\subsection{Optimization for gr-FCMSC}
Algorithm 1 can be generalized to optimize the problem of (\ref{grFCMSC_ObjectiveFunction}) in this section, and following ALM problem is constructed:
\begin{equation}
\begin{array}{l}
{\cal L}({E_x},{E_z},J,C,{Y_1},{Y_2},{Y_3})\\
 = {\left\| {{E_x}} \right\|_{2,1}} + {\lambda _1}{\left\| {{E_z}} \right\|_{2,1}}\\
 + {\lambda _2}{\left\| J \right\|_ * } + {\lambda _3}\sum\limits_{i = 1}^v {Tr({C^T}{L_i}C)} \\
 + \left\langle {{Y_1},X - XZ - {E_x}} \right\rangle  + \frac{\mu }{2}\left\| {X - XZ - {E_x}} \right\|_F^2\\
 + \left\langle {{Y_2},Z - ZC - {E_z}} \right\rangle  + \frac{\mu }{2}\left\| {Z - ZVC - {E_z}} \right\|_F^2\\
 + \left\langle {{Y_3},C - J} \right\rangle  + \frac{\mu }{2}\left\| {C - J} \right\|_F^2.
\end{array}
\end{equation}

Clearly, the subproblem with respect to $C$, which can be formulated as follows, is different from (\ref{FCMSC_Subproblem_C}):
\begin{equation}
\begin{array}{l}
\mathop {\min }\limits_C \left\langle {{Y_2},Z - ZC - {E_z}} \right\rangle  + \frac{\mu }{2}\left\| {Z - ZC - {E_z}} \right\|_F^2\\
 + \left\langle {{Y_3},C - J} \right\rangle  + \frac{\mu }{2}\left\| {C - J} \right\|_F^2{\rm{ + }}{\lambda _3}\sum\limits_{i = 1}^v {Tr({C^T}{L_i}C)} 
\end{array}
\end{equation}

And the optimization of the above problem is
\begin{equation}
\begin{array}{l}
C = T_{CA}^{ - 1}{T_{CB}},\\
{T_{CA}} = {\lambda _3}\sum\limits_{i = 1}^v {(L_i^T + {L_i})}  + \mu (I + {Z^T}Z),\\
{T_{CB}} = \mu J - {Y_3} + {Z^T}{Y_2} + \mu ({Z^T}Z - {Z^T}{E_z})
\end{array}
\end{equation}

As for other subproblems, we optimize they according to Algorithm 1 straightforward. And we skip they over for the compactness of this paper.


\subsection{Computational Complexity and Convergence}
As shown in Algorithm 1, the main computational burden is composed of five parts, i.e. the five corresponding subproblems. The complexity of updating $E_x$ is $O(dn^2+n^3)$, and the complexity of updating $E_z$ is $O(n^3)$, both of which are matrix multiplication. As for the subproblem of updating $J$, the complexity is $O(n^3)$. In the subproblem of updating $C$, the complexity is $O(n^3)$, since matrix inversion is included during optimization process. For updating $Z$, the Sylvester equation is optimized, and the complexity of this subproblem is $O(d^3+n^3)$. To sum up, the computational complexity of each iteration is $O(dn^2+d^3+n^3)$.

For the convergence analysis, unfortunately, we find that it is difficulty to give any solid proof on the convergence of the proposed algorithm, since more than two subproblems are involved during the optimization. Inspired by \cite{yin2015dual,zhang2017discriminativeBlockDiagonal,zhang2017discriminativeElasticNet}, convergence discussion will be presented in the experiments section, extensive experimental results on the real-world datasets show that the proposed algorithm can converge effectively with all-zero initialization except for variable $Z$, which is initialized with random values.

\section{Experiments}
In this section, extensive experiments are conducted on six benchmark datasets. Accordingly, experimental results are presented with the corresponding analyses. Both validation experiments and comparison experiments are provided, and the convergence properties and parameters sensitivity are analyzed as well. All codes are implemented in Matlab on a desktop with a four-core 3.6GHz processor and 8GB of memory.

\subsection{Experimental Settings}
To evaluate the performance of the proposed FCMSC, we employ six real-world datasets in experiments, including BBCSport\footnote{http://mlg.ucd.ie/datasets/} \cite{kumar2011co20,BBCSport}, Movies 617\footnote{http://lig-membres.imag.fr/grimal/data/movies617.tar.gz}, MSRCV1\footnote{http://research.microsoft.com/en-us/projects/objectclassrecognition/}, Olympics\footnote{http://mlg.ucd.ie/aggregation/}, ORL\footnote{https://www.cl.cam.ac.uk/research/dtg/attarchive/facedatabase.html}, and Yale Face\footnote{http://cvc.cs.yale.edu/cvc/projects/yalefaces/yalefaces.html}. {\color{black}{To be specific, BBCSport is collected from the BBC Sport website corresponding to sports news in 5 topical areas, and it consists of 544 documents, each which is divided into two sub-parts as two different views, and the standard TF-IDF normalization is utilized to obtain the corresponding features. Movie 617 is a movie dataset containing 617 movies of 17 genres, and it consists of two views, including keywords-mapping (view 1) and actors-mapping (view 2). MSRCV1 used in this paper consists of 210 images of 7 object classes, including building, cow, car airplane, tree, face, and bicycle, and 6 types of features are utilized, including: CENT (view1), CMT (view2), GIST (view3), HOG (view4), LBP (view5), and SIFT (view6). Olympics consists of 464 London 2012 Summer Olympics players' information active on Twitter of 28 different sports, and 9 different views are provide, including followedby-dictionary (view 1), follows-dictionary (view 2), mentionedby-dictionary (view 3), mentions-dictionary (view 4), retweets-dictionary (view 5), retweetedby-dictionary (view 6), listmerged500-dictionary (view 7), lists500-dictionary (view 8), tweets500-dictionary (view 9). ORL, which contains 400 images from 40 individuals, and Yale Face, which consists of 165 images from 15 individuals, are both face image datasets, three types of features, i.e., intensity (view1), LBP (view2) and Gabor (view3), are employed in datasets}}.

\begin{figure*}
\center
\includegraphics[width=1\linewidth]{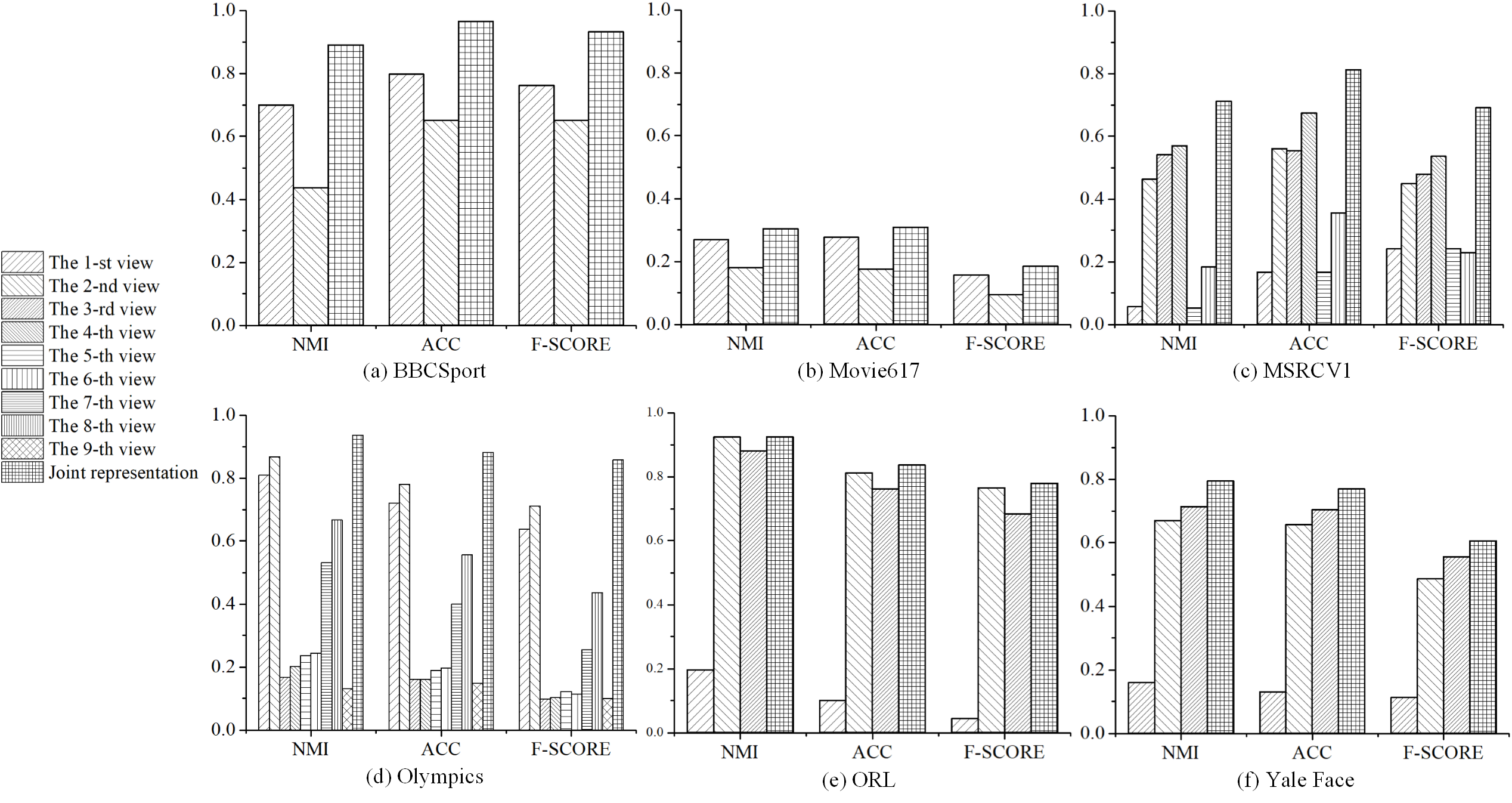}
\caption{Clustering results comparison between the proposed FCMSC conducted on the joint view representation and LRR performed on each single view. Six benchmark datasets are employed and clustering results are presented in the metric of NMI, ACC, and F-Score.}
\center
\label{Section5_Validation}
\end{figure*}

Meanwhile, three metrics are employed in this section to evaluate the clustering performance, including NMI (Normalized Mutual Information), ACC (accuracy), and F-score, which are commonly used in multi-view clustering \cite{GLMSC,brbic2018MLRSSC}. {\color{black}{To make more precise statements, the definition of the ACC utilized in this paper can be written as follows:}}
\begin{equation}
{\rm{ACC}} = \frac{{\sum\nolimits_{i = 1}^n {\sigma ({\tau _i},{\mathop{\rm map}\nolimits} ({\omega _i}))} }}{n},
\end{equation}
{\color{black}{where $x_i$ denotes the $i$-th sample, $\omega_i$ is the clustering label of the $i$-th sample, and $\tau_i$ illustrates the corresponding ground-truth label. The function of map$(\omega_i)$ is the permutation map function, in which the Kuhn-Munkres algorithm is employed. And $\sigma ( \cdot , \cdot )$ denotes the Dirac delta function.}} It should be noted that the higher value of all metrics corresponds the better clustering performance. All parameters of the competed methods are fine-tuned. To eliminate the randomness, 30 Monte Carlo (MC) trials are conducted with respect to each benchmark dataset. Experimental results are reported in form of the mean value and the standard deviation, and the best and the second best clustering results are present in bold font.

\subsection{Validation Experiments}
To validate our method, we compare the clustering results, achieved by the proposed FCMSC, with the clustering results, obtained by performing LRR on each single view. Specifically, validation experiments are conducted on all six benchmark datasets, and we shown the clustering performance of our methods and LRR on each single view with respect to NMI, ACC and F-score.

As shown in Fig. \ref{Section5_Validation}, the clustering performance of our FCMSC based on the joint view representation is much better than those of all single view. Taking BBCSport as example, NMI and ACC obtained by LRR \cite{liu2013robust11} with the best single view are $69.96\%$ and $79.70\%$ respectively. As for the proposed FCMSC based on the concatenated features, NMI and ACC are $89.04\%$ and $96.51\%$ respectively. In other words, our FCMSC achieves a relative increase of $27.27\%$ and $21.09\%$ with respect to NMI and ACC. Since FCMSC performs clustering on multiple view simultaneously and handles the cluster-specific corruptions properly, it can take advantage of consensus information to improve clustering results. Therefore, the proposed FCMSC is valid and can achieve promising clustering performance for multi-view data.

\subsection{Comparison Experiments}

\begin{figure*}
\center
\includegraphics[width=1\linewidth]{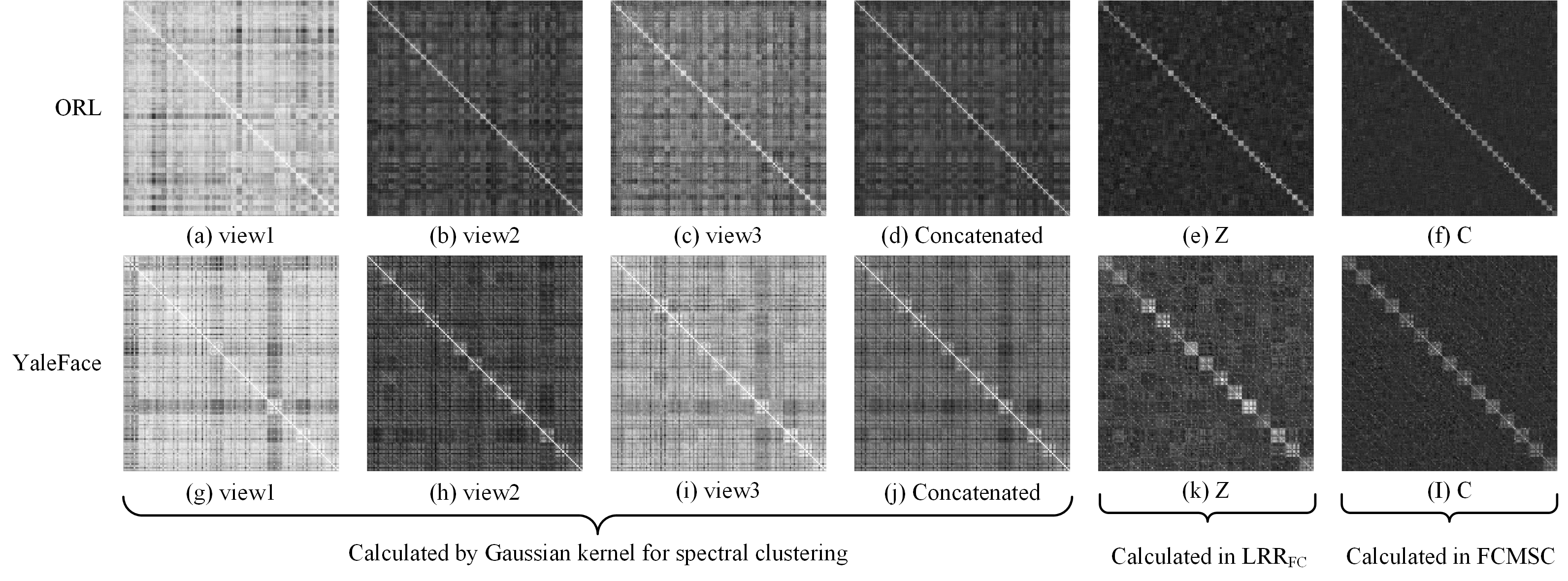}
\caption{Visualization of adjacency matrices (a, b, c, d, g, h, i, and j) and coefficient matrices (e, f, k, and l), where $Z$ and $C$ are both displayed in the form of (abs($Z^T$)+abs($Z$))/2 and (abs($C^T$)+abs($C$))/2.}
\label{Section5_Affinity_Matrix}
\center
\end{figure*}

To demonstrate the competitiveness of our FCMSC and gr-FCMSC, eleven approaches are employed for comparison and listed as follows:
\begin{itemize}
  \item [1)] 
  ${\rm SC_{BSV}}$ \cite{Luxburg2007A37}: Spectral Clustering of the Best Single View. Spectral clustering algorithm is employed on each single view, and the best clustering results of these views are presented.       
  \item [2)]
  ${\rm SC_{FC}}$: Spectral Clustering based on the Concatenated Features. Features of multiple views are concatenated, and then spectral clustering algorithm is applied to the joint view representation. 
  \item [3)]
  ${\rm LRR_{BSV}}$ \cite{liu2013robust11}: Low-Rank Representation of the Best Single View. Similar to the ${\rm SC_{BSV}}$, low-rank representation algorithm is conducted on each view, and the results of the view with the best clustering performance are reported.
  \item [4)]
  ${\rm LRR_{FC}}$: Low-Rank Representation based on the Concatenated Features. We apply the low-rank representation algorithm to the joint view representation to get multi-view clustering results.
  \item [5)]
  Kernel Addition \cite{kernelAddition}: This approach combines information of multi-view data by averaging the sum of kernel matrices of all views, then gets clustering results based on spectral clustering.
  \item [6)]
  Co-reg \cite{kumar2011co21}: Co-regularized multi-view spectral clustering. This approach clusters multi-view data by pursuing graph similarity agreement of multiple views.
  \item [7)]
  RMSC \cite{xia2014robust22}: Robust Multi-View Spectral Clustering via Low-Rank and Sparse Decomposition. RMSC recovers a common low-rank transition probability matrix via low-rank and sparse decomposition, and then obtain clustering results based on the standard Markov chains.
  \item [8)]
  LMSC \cite{zhang2017latent15}: Latent Multi-view Subspace Clustering. It learns a latent multi-view representation and attains the corresponding subspace coefficient matrix simultaneously.
  \item [9)]
  MLRSSC \cite{brbic2018MLRSSC}: Multi-view low-rank sparse subspace clustering. It learns the affinity matrix of multi-view data with the low-rank and sparsity constraints. Linear kernel MLRSSC algorithm is employed here for comparison.
  \item [10)]
  {\color{black}{DCCA \cite{DCCA}: Deep Canonical Correlation Analysis. It is a learning method which employs the neural networks to extend Canonical Correlation Analysis. We use k-means to obtain cluster results.}}
  \item [11)]
  {\color{black}{DCCAE \cite{DCCAE}: Deep Canonically Correlated Autoencoders. Two autoencoders are employed and DCCAE tries to maximize the canonical correlation between two views. K-means is utilized for clustering.}}
\end{itemize}

\begin{figure}
\center
\includegraphics[width=3.3in]{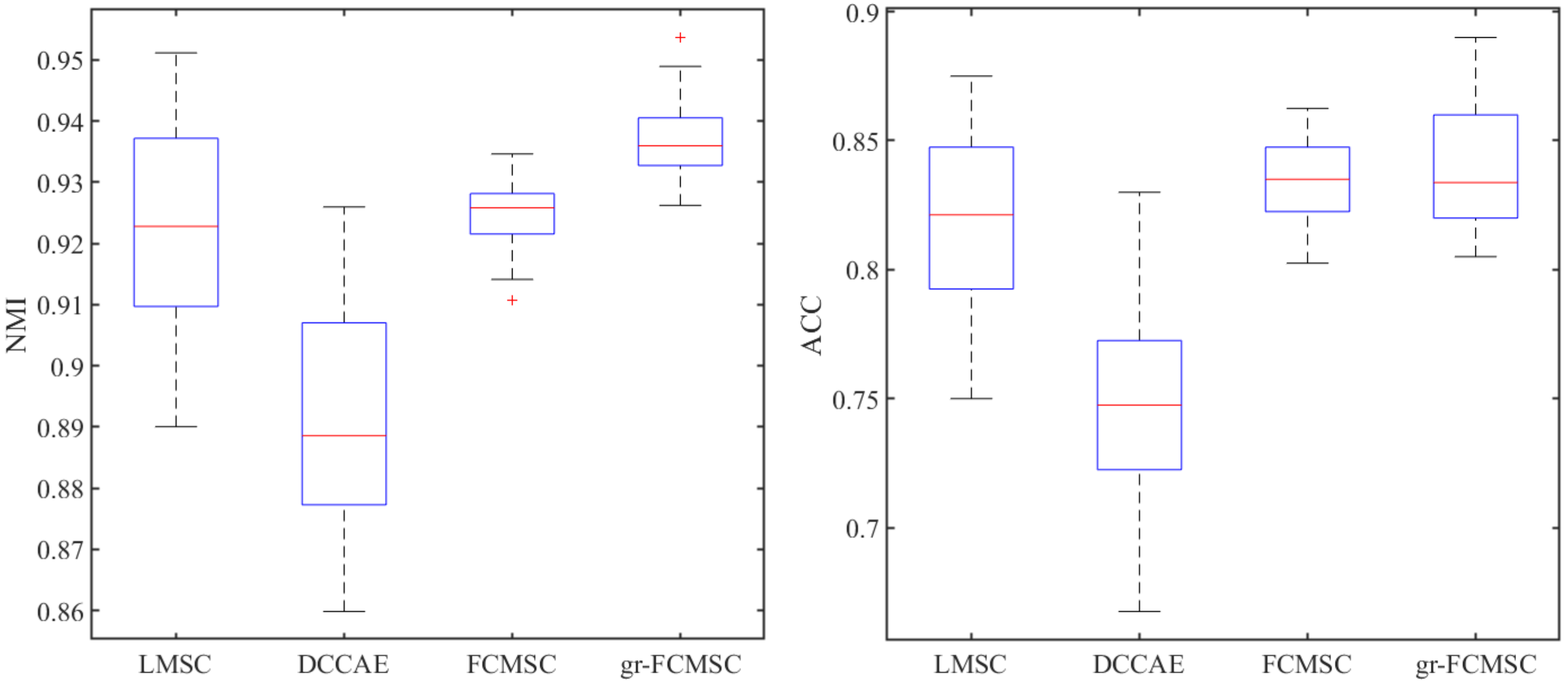}
\caption{Clustering results on ORL in metrics of NMI (left column) and ACC (right column).}
\center
\label{Section5_boxplot}
\end{figure}

As shown in Table \ref{tab:table_result} and \ref{tab:table_result_1}, experimental results are reported in form of the mean score, as well as the standard deviation. Overall, the proposed FCMSC and gr-FCMSC can achieve better clustering results on six benchmark datasets than other competed multi-view clustering approaches with respect to all the clustering metrics. For example, on MSRCV1 dataset with six views, our FCMSC gains a relative increase of 6.01\%, 4.48\% and 5.92\% with respect to NMI, ACC, and F-Score, respectively, comparing with the corresponding best competed method. Moreover, the proposed gr-FCMSC get 13.74\%, 10.69\%, and 16.03\% {\color{black}{relative improvement}} in metrics of NMI, ACC, and F-Score as well. {\color{black}{Furthermore, take experiments on ORL for examples, some statistical experimental results on ORL dataset are presented in Fig. \ref{Section5_boxplot}, it can be observed clearly that the proposed FCMSC and gr-FCMSC can achieve significantly improvement compared with competitors. Compared with deep learning based methods, i.e., DCCA and DCCAE, our methods can also achieve the better clustering performance, since both the sample-specific and the cluster-specific corruptions are taken into account for clustering.}}

Compared with ${\rm LRR_{FC}}$, the proposed FCMSC takes the cluster-specific corruptions, which are brought by multiple views, into consideration, and the clustering results indicate that it is essential to handle the clustering-specific corruptions during multi-view clustering. Since each view has its own specific properties that may be contrary to other views, it is difficult to explore and utilize the consensus information of multi-view data by performing some existing single-view clustering approaches on the concatenated features. To get an intuitive analysis, taking experiments on ORL and Yale Face datasets for example, Fig. \ref{Section5_Affinity_Matrix} presents the visualization of adjacency matrices, which are calculated from each view and concatenated features by different methods. Clearly, the adjacency matrix $C$ achieved by our proposed FCMSC has more suitable underlying structures for clustering.

\begin{table*}
\renewcommand{\arraystretch}{1}
\caption{Comparison results of different methods on the benchmark datasets}
\label{tab:table_result}
\centering
\setlength{\tabcolsep}{6mm}
\begin{tabular}{ccccc}
\toprule
\bfseries Dataset & \bfseries Method & \bfseries NMI  & \bfseries ACC & \bfseries F-SCORE\\
\midrule
\multirow{13}{*}{BBCSport}
& ${\rm SC_{BSV}}$ & 0.7182 (0.0054) & 0.8456 (0.0099) & 0.7671 (0.0067)\\
& ${\rm SC_{FC}}$ & 0.8019 (0.0095) & 0.8505 (0.0262) & 0.8452 (0.0206)\\
& ${\rm LRR_{BSV}}$ & 0.6996 (0.0001) & 0.7970 (0.0015) & 0.7612 (0.0001)\\
& ${\rm LRR_{FC}}$ & 0.5580 (0.0110) & 0.6684 (0.0088) & 0.6064 (0.0055)\\
& Kernel Addition & 0.6170 (0.0085) & 0.7347 (0.0099) & 0.6684 (0.0059)\\
& Co-reg & 0.7185 (0.0031) & 0.8465 (0.0050) & 0.7674 (0.0041)\\
& RMSC & 0.8124 (0.0074) & 0.8562 (0.0198) & 0.8514 (0.0132)\\
& LMSC & 0.8393 (0.0043) & 0.9180 (0.0031) & 0.8996 (0.0033)\\
& MLRSSC & 0.8855 (0.0000) & \textbf{0.9651 (0.0000)} & 0.9296 (0.0000)\\
& {\color{black}{DCCA}} & {\color{black}{0.2779 (0.0041)}} & {\color{black}{0.5490 (0.0042)}} & {\color{black}{0.4116 (0.0010)}}\\
& {\color{black}{DCCAE}} & {\color{black}{0.3298 (0.0046)}} & {\color{black}{0.5438 (0.0089)}} & {\color{black}{0.4088 (0.0080)}}\\
& FCMSC & \textbf{0.8904 (0.0000)} & \textbf{0.9651 (0.0000)} & \textbf{0.9317 (0.0000)}\\
& gr-FCMSC & \textbf{0.8973 (0.0000)} & \textbf{0.9670 (0.0000)} & \textbf{0.9348 (0.0000)}\\
\hline
\multirow{13}{*}{Movies 617}
& ${\rm SC_{BSV}}$ & 0.2606 (0.0020) & 0.2579 (0.0035) & 0.1481 (0.0025)\\
& ${\rm SC_{FC}}$ & 0.2668 (0.0017) & 0.2604 (0.0033) & 0.1542 (0.0019)\\
& ${\rm LRR_{BSV}}$ & 0.2667 (0.0059) & 0.2747 (0.0071) & 0.1545 (0.0047)\\
& ${\rm LRR_{FC}}$ & 0.2839 (0.0075) & 0.2824 (0.0135) & 0.1813 (0.0063)\\
& Kernel Addition & 0.2917 (0.0026) & 0.2901 (0.0049) & 0.1764 (0.0033)\\
& Co-reg & 0.2454 (0.0018) & 0.2396 (0.0017) & 0.1381 (0.0016)\\
& RMSC & 0.2957 (0.0032) & 0.2971 (0.0040) & 0.1810 (0.0028)\\
& LMSC & 0.2813 (0.0098) & 0.2747 (0.0094) & 0.1606 (0.0068)\\
& MLRSSC & 0.2975 (0.0061) & 0.2887 (0.0111) & 0.1766 (0.0068)\\
& {\color{black}{DCCA}} & {\color{black}{0.1764 (0.0002)}} & {\color{black}{0.1948 (0.0018)}} & {\color{black}{0.1141 (0.0014)}}\\
& {\color{black}{DCCAE}} & {\color{black}{0.1759 (0.0061)}} & {\color{black}{0.2009 (0.0118)}} & {\color{black}{0.1214 (0.0056)}}\\
& FCMSC & \textbf{0.3043 (0.0052)} & \textbf{0.3090 (0.0063)} & \textbf{0.1852 (0.0034)}\\
& gr-FCMSC & \textbf{0.3169 (0.0059)} & \textbf{0.3051 (0.0049)} & \textbf{0.1930 (0.0035)}\\
\hline
\multirow{13}{*}{MSRCV1}
& ${\rm SC_{BSV}}$ & 0.6047 (0.0112) & 0.6826 (0.0171) & 0.5724 (0.0122)\\
& ${\rm SC_{FC}}$ & 0.4398 (0.0021) & 0.5073 (0.0077) & 0.3978 (0.0032)\\
& ${\rm LRR_{BSV}}$ & 0.5704 (0.0054) & 0.6732 (0.0091) & 0.5368 (0.0076)\\
& ${\rm LRR_{FC}}$ & 0.6257 (0.0105)	& 0.6871 (0.0105) & 0.5913 (0.0142)\\
& Kernel Addition & 0.6176 (0.0087) & 0.7102 (0.0130) & 0.5973 (0.0097)\\
& Co-reg & 0.6583 (0.0106) & 0.7674 (0.0169)& 0.6459 (0.0128)\\
& RMSC & 0.6696 (0.0064) & 0.7819 (0.0125) & 0.6614 (0.0093)\\
& LMSC & 0.6162 (0.0676) & 0.6992 (0.0700) & 0.5936 (0.0763)\\
& MLRSSC & 0.6709 (0.0352) & 0.7774 (0.0497) & 0.6524 (0.0470)\\
& {\color{black}{DCCA}} & {\color{black}{0.6606 (0.0000)}} & {\color{black}{0.7429 (0.0000)}} & {\color{black}{0.6270 (0.0000)}}\\
& {\color{black}{DCCAE}} & {\color{black}{0.6782 (0.0040)}} & {\color{black}{0.7662 (0.0015)}} & {\color{black}{0.6462 (0.0025)}}\\
& FCMSC & \textbf{0.7112 (0.0031)} & \textbf{0.8122 (0.0030)} & \textbf{0.6910 (0.0046)}\\
& gr-FCMSC & \textbf{0.7631 (0.0036)} & \textbf{0.8605 (0.0022)} & \textbf{0.7570 (0.0036)}\\
\bottomrule
\end{tabular}
\end{table*}

\begin{table*}
\renewcommand{\arraystretch}{1}
\caption{Comparison results of different methods on the benchmark datasets}
\label{tab:table_result_1}
\centering
\setlength{\tabcolsep}{6mm}
\begin{tabular}{ccccc}
\toprule
\bfseries Dataset & \bfseries Method & \bfseries NMI  & \bfseries ACC & \bfseries F-SCORE\\
\midrule
\multirow{13}{*}{Olympics}
& ${\rm SC_{BSV}}$ & 0.7617 (0.0046) & 0.6288 (0.0112) & 0.5178 (0.0134)\\
& ${\rm SC_{FC}}$ & 0.5625 (0.0038) & 0.4610 (0.0078) & 0.3194 (0.0085)\\
& ${\rm LRR_{BSV}}$ & 0.8674 (0.0038) & 0.7830 (0.0093) & 0.7112 (0.0088)\\
& ${\rm LRR_{FC}}$ & 0.8910 (0.0054) & 0.7746 (0.0190) & 0.7532 (0.0190)\\
& Kernel Addition & 0.7245 (0.0038) & 0.6093 (0.0073) & 0.5189 (0.0071)\\
& Co-reg & 0.8308 (0.0027) & 0.7341 (0.0071) & 0.6707 (0.0079)\\
& RMSC & 0.7573 (0.0063) & 0.6372 (0.0108) & 0.5687 (0.0117)\\
& LMSC & 0.8902 (0.0065) & 0.8043 (0.0140) & 0.7814 (0.0154)\\
& MLRSSC & 0.9122 (0.0067) & 0.8454 (0.0208) & 0.8236 (0.0285)\\
& {\color{black}{DCCA}} & {\color{black}{0.7782 (0.0015)}} & {\color{black}{0.6474 (0.0047)}} & {\color{black}{0.4024 (0.0072)}}\\
& {\color{black}{DCCAE}} & {\color{black}{0.7686 (0.0042)}} & {\color{black}{0.6515 (0.0115)}} & {\color{black}{0.4096 (0.0172)}}\\
& FCMSC & \textbf{0.9357 (0.0062)} & \textbf{0.8815 (0.0199)} & \textbf{0.8576 (0.0264)}\\
& gr-FCMSC & \textbf{0.9389 (0.0037)} & \textbf{0.8890 (0.0137)} & \textbf{0.8649 (0.0176)}\\
\hline
\multirow{13}{*}{ORL}
& ${\rm SC_{BSV}}$ & 0.8868 (0.0069) & 0.7459 (0.0121) & 0.6805 (0.0159)\\
& ${\rm SC_{FC}}$ & 0.8084 (0.0027) & 0.6323 (0.0061) & 0.5236 (0.0069)\\
& ${\rm LRR_{BSV}}$ & 0.9240 (0.0054) & 0.8122 (0.0203) & 0.7650 (0.0166)\\
& ${\rm LRR_{FC}}$ & 0.8497 (0.0085) & 0.7178 (0.0190) & 0.6119 (0.0218)\\
& Kernel Addition & 0.8028 (0.0033) & 0.6349 (0.0074) & 0.5224 (0.0061)\\
& Co-reg & 0.8277 (0.0040) & 0.6653 (0.0080) & 0.5672 (0.0092)\\
& RMSC & 0.8885 (0.0056) & 0.7482 (0.0128) & 0.6866 (0.0139)\\
& LMSC & 0.9215 (0.0168) & 0.8193 (0.0360) & 0.7623 (0.0419)\\
& MLRSSC & 0.9102 (0.0113) & 0.8042 (0.0234) & 0.7459 (0.0281)\\
& {\color{black}{DCCA}} & {\color{black}{0.8589 (0.0025)}} & {\color{black}{0.7093 (0.0041)}} & {\color{black}{0.5716 (0.0001)}}\\
& {\color{black}{DCCAE}} & {\color{black}{0.8906 (0.0171)}} & {\color{black}{0.7476 (0.0356)}} & {\color{black}{0.6741 (0.0399)}}\\
& FCMSC & \textbf{0.9249 (0.0055)} & \textbf{0.8359 (0.0165)} & \textbf{0.7792 (0.0180)}\\
& gr-FCMSC & \textbf{0.9370 (0.0065)} & \textbf{0.8382 (0.0229)} & \textbf{0.7991 (0.0227)}\\
\hline
\multirow{13}{*}{Yale Face}
& ${\rm SC_{BSV}}$ & 0.6229 (0.0354) & 0.5715 (0.0497) & 0.4319 (0.0472)\\
& ${\rm SC_{FC}}$ & 0.5761 (0.0335) & 0.5145 (0.0460) & 0.3653 (0.0420)\\
& ${\rm LRR_{BSV}}$ & 0.7134 (0.0098) & 0.7034 (0.0125) & 0.5561 (0.0159)\\
& ${\rm LRR_{FC}}$ & 0.6917 (0.0190) & 0.6667 (0.0236) & 0.4941 (0.0303)\\
& Kernel Addition & 0.5872 (0.0320) & 0.5352 (0.0397) & 0.3823 (0.0390)\\
& Co-reg & 0.6146 (0.0084) & 0.5638 (0.0108) & 0.4208 (0.0110)\\
& RMSC & 0.6590 (0.0108) & 0.6091 (0.0161) & 0.4773 (0.0133)\\
& LMSC & 0.7073 (0.0105) & 0.6758 (0.0116) & 0.5138 (0.0172)\\
& MLRSSC & 0.7005 (0.0311) & 0.6733 (0.0384) & 0.5399 (0.0377)\\
& {\color{black}{DCCA}} & {\color{black}{0.7642 (0.0004)}} & {\color{black}{0.7392 (0.0011)}} & {\color{black}{0.6159 (0.0012)}}\\
& {\color{black}{DCCAE}} & {\color{black}{0.6888 (0.0226)}} & {\color{black}{0.6442 (0.0291)}} & {\color{black}{0.5152 (0.0307)}}\\
& FCMSC & \textbf{0.7939 (0.0206)} & \textbf{0.7691 (0.0267)} & \textbf{0.6058 (0.0306)}\\
& gr-FCMSC & \textbf{0.7979 (0.0202)} & \textbf{0.7717 (0.0222)} & \textbf{0.6095 (0.0335)}\\
\bottomrule
\end{tabular}
\end{table*}

\begin{figure*}
\center
\includegraphics[width=1\linewidth]{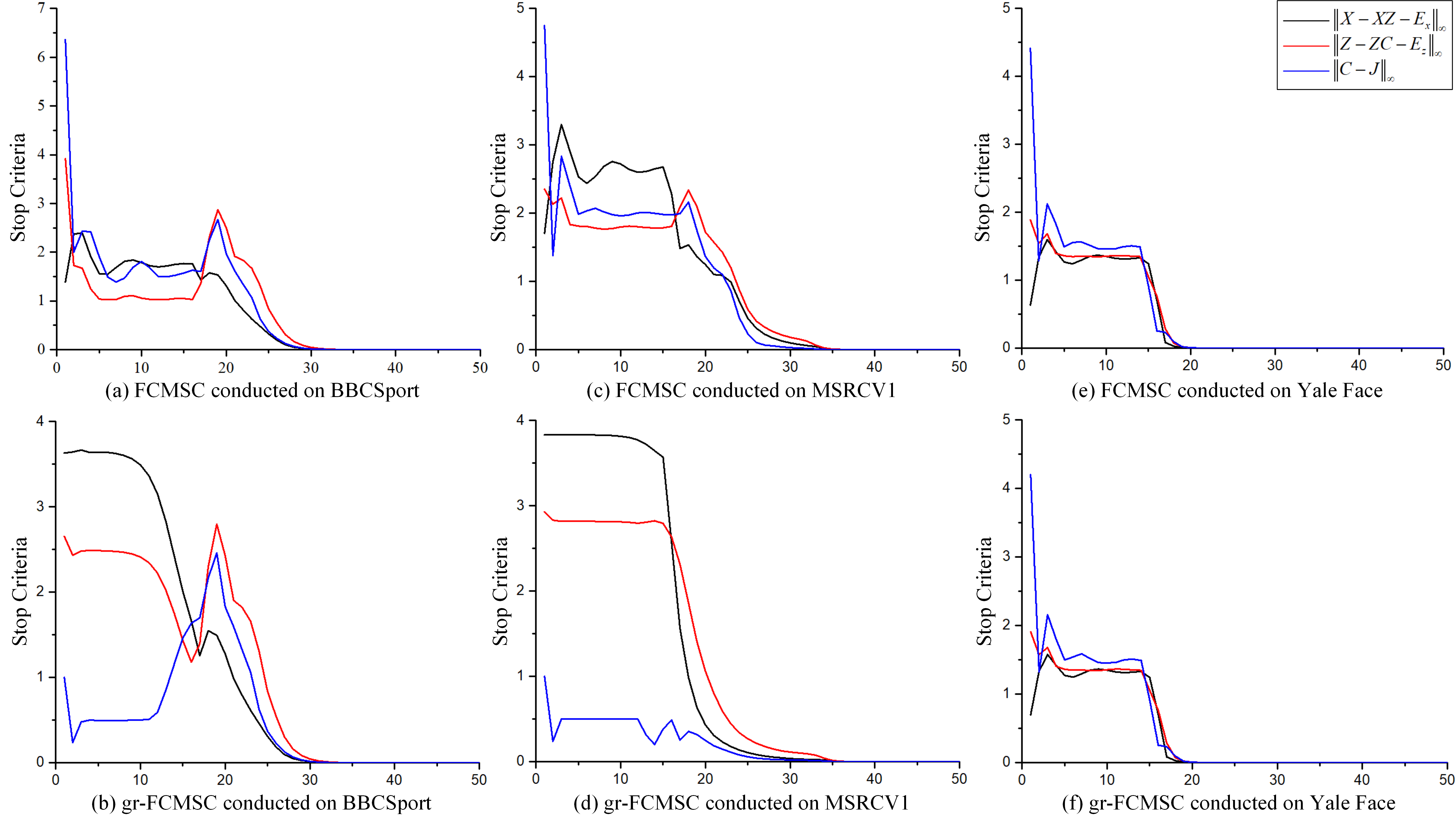}
\caption{Convergence of the proposed FCMSC and gr-FCMSC. Convergence curves about the stop criteria of reconstruction errors versus the iteration numbers on three datasets, including (a)-(b) BBCSport, (c)-(d) MSRCV1, and (e)-(f) Yale Face, are displayed in this section.}
\center
\label{Section5_ConvAna}
\end{figure*}

\subsection{Comparisons Between FCMSC and gr-FCMSC}
Both the proposed FCMSC and gr-FCMSC perform clustering on multiple views simultaneously, and get promising clustering results. The consensus information of multi-view data is both leveraged by FCMSC and gr-FCMSC. Additionally, gr-FCMSC also takes advantage of the complementary of multiple views by means of graph Laplacian regularizers for multi-view clustering. As shown in Table \ref{tab:table_result}, gr-FCMSC improves FCMSC significantly by exploring the complementary information. For example, Compared with FCMSC, gr-FCMSC gains 7.30\% and 5.95\% relative improvement on MSRCV1 in metrics of NMI and ACC. And on Movie 617 dataset, gr-FCMSC achieves 3.42\% and 4.04\% relative improvement with respect to the metrics of NMI and F-Score.

\subsection{Parameters Sensitivity and Convergence Analysis}

\begin{figure}
\center
\includegraphics[width=3.3in]{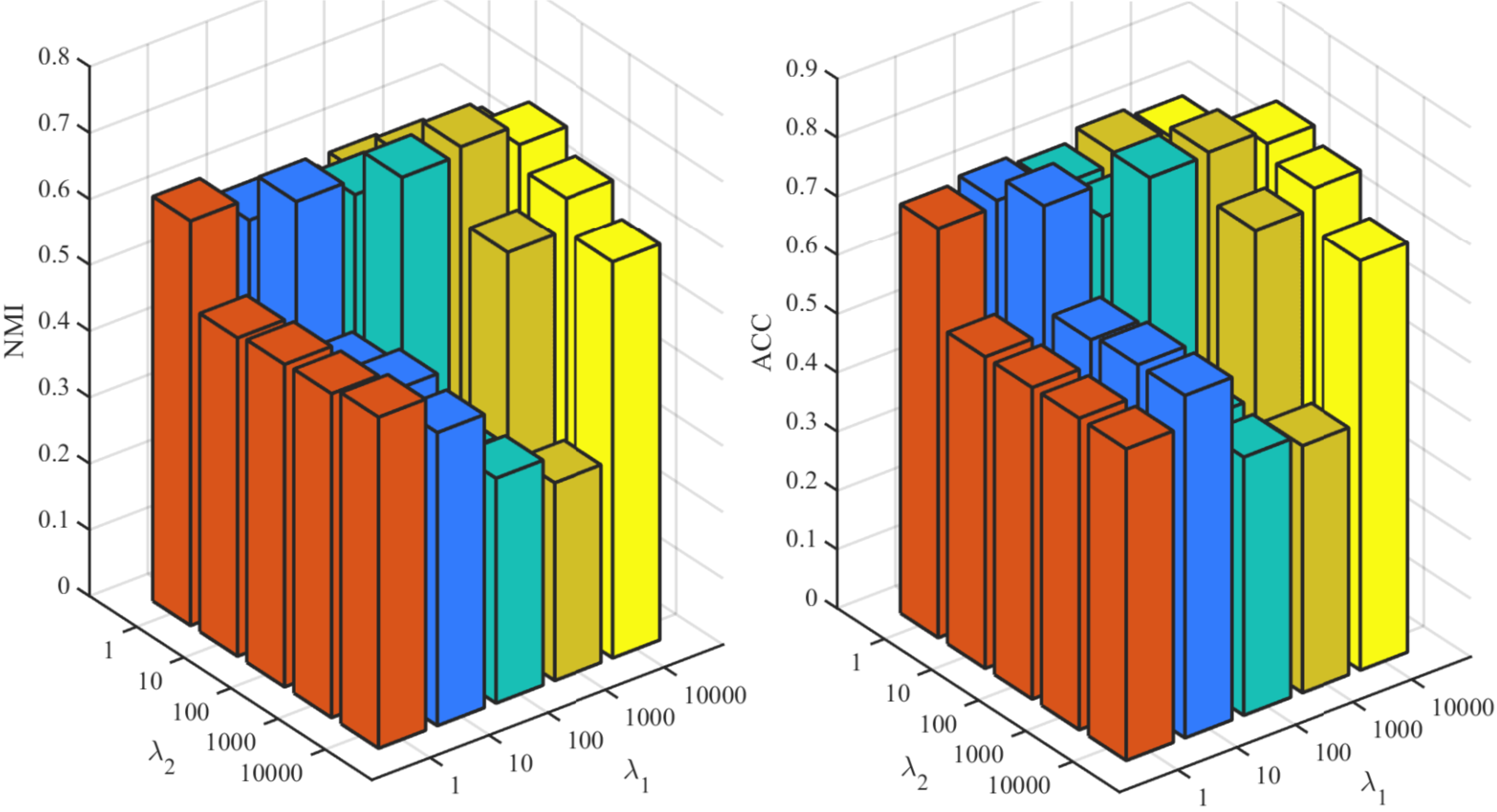}
\caption{Clustering results of proposed FCMSC with different $\lambda _1$ and $\lambda _2$ on MSRCV1 dataset.}
\center
\label{Section5_lambda1_2}
\end{figure}

\begin{figure}
\center
\includegraphics[width=3.3in]{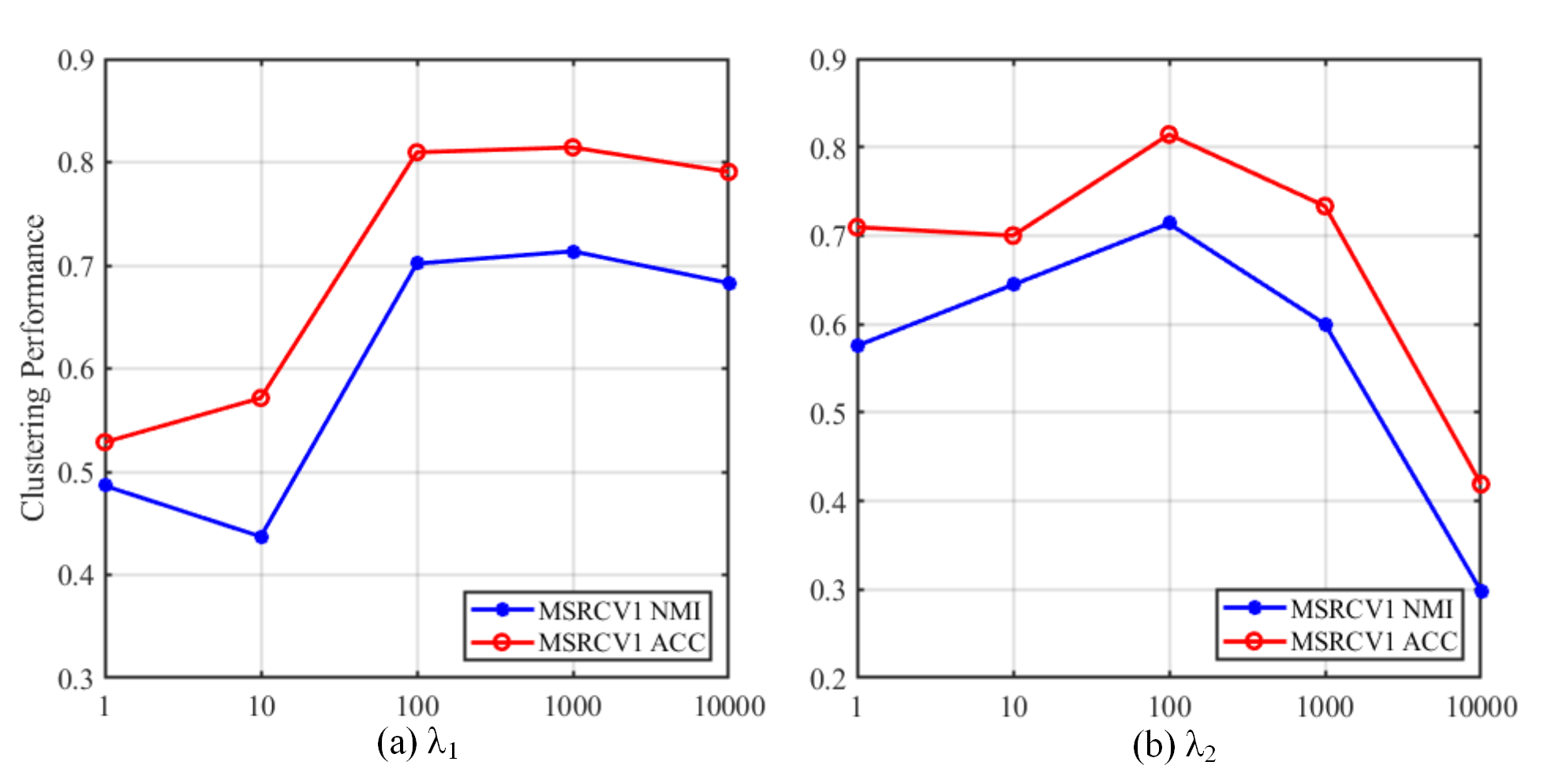}
\caption{Influence of $\lambda _1$ and $\lambda _2$ on MSRCV1 dataset. (a) illustrates the influence of $\lambda _1$ with $\lambda _2=100$, (b) illustrates the influence of $\lambda _2$ with $\lambda _1=1000$.}
\center
\label{Section5_lambda1_2_separate}
\end{figure}

\begin{figure}
\center
\includegraphics[width=3.3in]{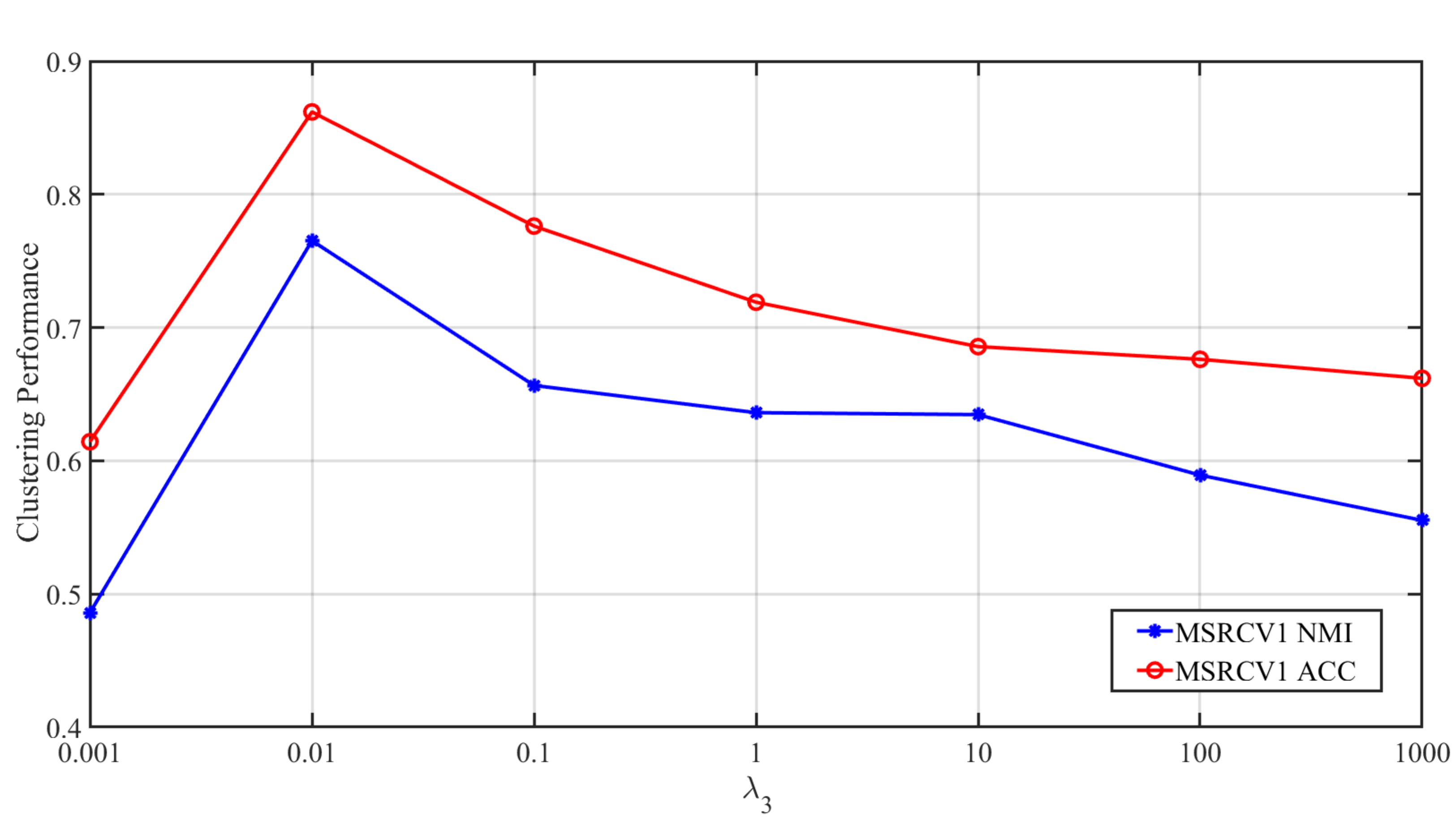}
\caption{Clustering results of proposed gr-FCMSC with different $\lambda _3$ on MSRCV1 dataset.}
\center
\label{Section5_lambda_3}
\end{figure}

Besides, Convergence analysis, shown in Fig. \ref{Section5_ConvAna}, and parameters influence of the proposed methods, shown in Fig. \ref{Section5_lambda1_2}, Fig. \ref{Section5_lambda1_2_separate}, and Fig. \ref{Section5_lambda_3}, are discussed in the this section as well.

In the proposed methods, {\color{black}{as shown in the objective functions (\ref{FCMSC_ObjectiveFunction}) and (\ref{grFCMSC_ObjectiveFunction}), there are two trade-off parameters required to be fine-tuned, i.e. ${\lambda _1}$ and ${\lambda _2}$, for FCMSC, and an extra trade-off parameter ${\lambda _3}$ for gr-FCMSC. Values of ${\lambda _1}$ and ${\lambda _2}$ are selected from $\{1, 10, 100, 1000, 10000\}$. As shown in Fig. \ref{Section5_lambda1_2} and Fig. \ref{Section5_lambda1_2_separate}, it can be observed that promising clustering results can be attained when ${\lambda _1}$ is relative large and $\lambda _2$ equals to 100. To be specific, curves drawn in Fig. 8 demonstrate the influence of ${\lambda _1}$ and ${\lambda _2}$, respectively. For ${\lambda _1}$, it can be observed that promising clustering results can be achieved with a relatively large value. As for ${\lambda _2}$, generally speaking, the prior knowledge of the dataset error level determines the choice of ${\lambda _2}$ mainly, and the promising clustering performance can be achieved with ${\lambda _2}=100$ on MSRCV1 dataset. Meanwhile, the influence of ${\lambda _3}$ for gr-FCMSC is also discussed as shown in Fig. \ref{Section5_lambda1_2_separate}, it can be observed that the best clustering performance can be attained with ${\lambda _3}=0.01$, and the clustering performance degenerates when the value of ${\lambda _3}$ is larger than $0.01$.}}

Meanwhile, we explore the convergence properties of the proposed FCMSC and gr-FCMSC. Fig. \ref{Section5_ConvAna} displays the convergence of our approaches conducted on three datasets, including BBCSport, MSRCV1, and Yale Face. It can be observed that both FCMSC and gr-FCMSC can achieve the quick convergence within 40 iterations. Although it is difficult for us to give an solid proof on the convergence, experimental results demonstrate the effectiveness and convergence of our methods empirically.


\section{Conclusion}
This paper proposes a feature concatenation multi-view subspace clustering approach, termed FCMSC, and a graph regularized FCMSC (gr-FCMSC) as well. Different from most of existing approaches, the proposed methods can perform clustering on all views simultaneously by exploring the consensus information and complementary information of multi-view data based on the concatenated features. By taking the cluster-specific corruptions into consideration, the proposed methods can obtain a desired coefficient matrix and achieve promising clustering results. Extensive experiments on six benchmark datasets demonstrate the superiority of our approach over some state-of-the-arts.

Despite effectiveness of the proposed methods, they are time consuming due to the operation of matrix inversion and SVD decomposition involved in the optimization, especially when the number of data is large. Further work will focus on the improvement of proposed methods for large-scale data, by employing the dimensionality reduction and the binary representation \cite{Zhang2018Binary39} strategies. {\color{black}{And for the gr-FCMSC, the incompatible graph information among multiple views is ignored, further work will also focus on this problem.}}

\section*{Acknowledgements}

This work is supported by the National Natural Science Foundation of China under Grant No.
61573273.

\printcredits

\bibliographystyle{cas-model2-names}

\bibliography{FCMSC}

\end{document}